\pdfoutput=1

\documentclass[11pt]{article}

\usepackage[preprint]{acl}

\usepackage{times}
\usepackage{latexsym}
\usepackage{caption}

\usepackage[T1]{fontenc}

\usepackage[utf8]{inputenc}

\usepackage{microtype}

\usepackage{inconsolata}

\usepackage{graphicx}
\usepackage{placeins}
\usepackage[utf8]{inputenc} 
\usepackage[T1]{fontenc}    
\usepackage{hyperref}       
\usepackage{url}            
\usepackage{booktabs}       
\usepackage{amsfonts}       
\usepackage{nicefrac}       
\usepackage{microtype}      
\usepackage{xcolor}         
\usepackage[ruled,vlined,linesnumbered]{algorithm2e}
\usepackage{amssymb,amsmath,amsthm}
\usepackage{booktabs}
\usepackage{upgreek}
\usepackage{graphicx}
\usepackage{caption}
\usepackage{subcaption}
\usepackage{multirow}
\usepackage{enumitem}
\usepackage{xcolor}
\usepackage{multirow}
\usepackage{stackengine}
\usepackage{mathtools}
\usepackage[hang,flushmargin]{footmisc}
\usepackage{comment}
\usepackage{subcaption}
\usepackage{natbib}
\setcitestyle{square, comma, numbers, super}

\newtheorem{lemma}{Lemma}

\title{Aligned at the Start: Conceptual Groupings in LLM Embeddings}

\author{Mehrdad Khatir \\
  Virginia Tech \\
  \texttt{khatir@vt.edu} \\\And
  Sanchit Kabra \\
  Virginia Tech\\
  \texttt{sanchit23@vt.edu} \\ \And
  Chandan K. Reddy \\
  Virginia Tech\\
  \texttt{reddy@cs.vt.edu}
  }

\begin{document}
\maketitle

\begin{abstract}
This paper shifts focus to the often-overlooked input embeddings – the initial representations fed into transformer blocks. Using fuzzy graph, k-nearest neighbor (k-NN), and community detection, we analyze embeddings from diverse LLMs, finding significant categorical community structure aligned with predefined concepts and categories aligned with humans. We observe these groupings exhibit within-cluster organization (such as hierarchies, topological ordering, etc.), hypothesizing a fundamental structure that precedes contextual processing. To further investigate the conceptual nature of these groupings, we explore cross-model alignments across different LLM categories within their input embeddings, observing a medium to high degree of alignment. Furthermore, provide evidence that manipulating these groupings can play a functional role in mitigating ethnicity bias in LLM tasks.
\end{abstract}

\section{Introduction}
Large Language models (LLMs) are rapidly approaching human-level language abilities. However, these advancements raise important concerns - despite their immense potential, language models are susceptible to unexpected and potentially harmful behaviors such as hallucination \citep{ji2023survey}, stereotyping \cite{pmlr-v139-liang21a}, misinformation \cite{pan-etal-2023-risk}, and leakage of sensitive training data \cite{10179300}. 
Moreover, the extent of their capabilities and the nature of their understanding is not yet known, thus capturing the attention of diverse groups, from policymakers to academics \cite{gu-etal-2023-language, peng2022copen, stochastic-parrot, gurnee2024language}. 
While these issues may seem orthogonal, they stem from a common concern - the "black box" nature of large language models. The limited ability to understand how they reach their outputs raises doubts about their true capabilities, potential biases, and unintended consequences.

A crucial step towards addressing these challenges lies in understanding how LLMs internally represent and manipulate semantic concepts, the building blocks of human language and thought. Unraveling these internal representations is essential for not only improving LLM performance and robustness, but also for building trustworthy machine learning systems.

In the realm of contextual representations, some studies \cite{patel2021mapping, gurnee2024language, abdou2021can} have investigated the ability of transformer-based models to learn representations of color, spatial, and temporal information. These studies often rely on analyzing contextual embeddings, which are intermediate outputs of LLMs. Moreover, contextual embeddings are inherently tied to the specific input context, making them highly variable, less generalizable, and less controllable/manipulatable. 

In our work, we explore this critical area by investigating the emergence of conceptual structures within the \textbf{input embedding space} of LLMs. We aim to explore if these conceptual formations form independent of the context, and if such groupings are in alignment across LLMs, if these clusters exhibit internal organization, and if these patterns have functional implications for model behavior.
Specifically, we examine:
\setlist{nolistsep}
\begin{itemize}[leftmargin=*] 
    \item RQ1: Whether semantically related words and phrases are grouped together, forming identifiable conceptual clusters that are aligned with external world concepts and categories. 
    \item RQ2: Do these groupings exhibit intra-cluster organization such as hierarchies, topological ordering, etc., thereby suggesting the formation of structured concepts. 
    \item RQ3: Is there an inter-model alignment in semantic organization, across diverse transformer-based LLMs (Albert \cite{lan2019albert}, T5 \cite{raffel2020exploring}, Llama3 \cite{meta_llama_3}) irrespective of architecture, size, or pretraining data. 
    \item RQ4: Whether these groupings play a functional role in LLMs.  We test this by a case study that tries to mitigate ethnicity bias through cluster modification. 
\end{itemize}

To uncover conceptual structures within the input embedding space of LLMs, we employ fuzzy graph construction \cite{mcinnes2020umap}. Then, the fuzzy graph is analyzed using a community detection algorithm \cite{blondel2008fast} to reveal conceptual groupings and their categorical organization. We use Louvain community detection, which is effective at revealing hierarchical community structures \cite{blondel2008fast} in conjunction with multiple k choices (for k-nearest neighbors) in our approach. This systemic approach allows us to investigate both the existence and the hierarchical organization of conceptual clusters, directly addressing our research questions.
Quantitative evaluation on external datasets (named entities \cite{NameDataset2021, CountryDataset2018}, numerical tokens, and social structures) demonstrates that token embeddings exhibit significant categorical community structure aligned with real-world concepts.

The structure of this paper is as follows: We first establish necessary background on embeddings, semantic representations, and evaluation strategies (Section \ref{prelim}), followed by a description of our methodological approach (Section \ref{con-extract}).  We then present our core findings, starting with LLM-human alignment (Section \ref{human-lm-results}), with a focus on within-cluster properties and hierarchical structure. Then, we explore LLM-LLM alignment within their input embedding space (Section \ref{llm-llm}). Section 6 demonstrates the practical implications of our work through embedding engineering and bias mitigation.  Finally, we conclude in Section \ref{conclusion}.

\section{Preliminaries}\label{prelim}

\subsection{Static, Contextual and Base Embeddings}
In this section, we clarify the distinctions between static, contextual, and base embeddings, which are crucial for understanding modern language models.
\begin{itemize}[leftmargin=*]
    \item \textbf{Static embeddings} (e.g., Word2Vec\cite{mikolov2013efficient}, GloVe \cite{pennington-etal-2014-glove}) are context-independent vector representations of words, meaning each word has a fixed embedding regardless of its surrounding text. This limits their ability to handle polysemy (words with multiple meanings). These embeddings are typically pre-trained on large corpora and can be used in various downstream tasks. Critically, for the context of this work, static embeddings are product of legacy LM models and not used as the input representations of modern transformer-based LLMs. They are not inputs to transformer blocks and thus have limited significance when it comes to applications such as mitigation techniques (e.g. embedding engineering) in LLMs. 
    \item \textbf{Contextual embeddings} These are dynamic, context-dependent vectors. The embedding of a token is a function of its surrounding text, enabling the representation of nuanced meaning and resolving polysemy (e.g., Bert \cite{devlin2018bert}, Albert \cite{lan2019albert}, GPT variants \cite{raffel2020exploring, meta_llama_3}). Different model layers capture different levels of contextual conditioning. However, this context-dependence limits direct interpretability and generalizability of individual token embeddings outside of specific contexts.
    \item \textbf{Base Embeddings}: The process of generating contextual embeddings starts with base embeddings, which provide the initial vector representation for each input token.  These differ from static embeddings such as GloVe and Word2Vec from the following perspectives. (1) Generation: Base embeddings are learned parameters within the LLM, as compared to separately trained static embeddings. (2) Usage: They are the direct input to the transformer blocks, forming the basis upon which contextualized representations are built through the model's subsequent layers. These embeddings are the focus of our study.
\end{itemize}

\subsection{Previous Works on Embedding Interpretability}
Previous research on interpretability in LLMs has primarily focused on analyzing either contextual embeddings (for modern LLMs) or static embeddings (for legacy language models which are not directly applicable to LLMs due to the architectural difference).

In the realm of contextual representations, initial research focused on the learning dynamics of linguistic features within LLMs \cite{tenney2018you, liu-etal-2019-linguistic}, the scope has expanded to explore how these models acquire and represent knowledge about the world. Some studies \cite{patel2021mapping, gurnee2024language, abdou2021can} have investigated the ability of transformer-based models to learn representations of color, spatial, and temporal information.
These studies often rely on analyzing contextual embeddings, which are intermediate outputs of LLMs. Moreover, contextual embeddings are inherently tied to the specific input context, making them highly variable and less generalizable.

The closest work to our knowledge that focuses on context agnostic embeddings within the modern LLMs is \cite{bommasani-etal-2020-interpreting} where the authors propose a method to create context agnostic word embeddings from contextualized word representations using (sub)-word pooling as well as context combination techniques, and tested on semantic similarity datasets. Furthermore, \cite{ijcai2021p0530} proposes a method for creating context-agnostic word representations by averaging the contextual embeddings derived from BERT, given a set of inferences on a masked target token within a corpus. In this methodology, the context is seen as a form of Gaussian noise that can be averaged out and hence produces a context-agnostic semantic representation. They observed such embedding represent richer semantic information than static word embedding counterparts (Word2Vec and GloVe) in intrinsic evaluation tasks.
However, these works lack rigorous and extensive analysis of base embeddings of the LLMs to explores the intrinsic semantic organization within input embeddings of LLMs, possbile conceptual groupings and their alignments.  

\section{Concept Extraction}\label{con-extract}
To investigate RQ1 and RQ2, we study the Human-LLM alignment of the input representations. To this end, we develop a method to first extract possible formed concepts within that space (refer to appendix \ref{concept-formation} for the discussion of conceptual groupings and semantic memory), and then to evaluate them against external datasets.

Our methodology consists of building the semantic graph, then using community detection to extract possible conceptual groupings.\footnote{Note that we favor community detection over traditional clustering for identifying the conceptual clusters due to (1) its ability to align with the network-like structure of semantic representations, (2) its independence from the need for a predetermined number of clusters, and (3) its effectiveness in managing high-dimensional data by transforming it into a graph.}
Note that differentiable embedding functions guarantee a smooth, optimizable embedding space, but do not ensure a uniform distribution of concept instances. This unevenness, arising from factors like varying concept complexity and instance frequency, implies that while conceptual groupings may be locally uniform, the overall embedding space can be unevenly distributed.  Therefore, effectively identifying these groupings requires mitigating this unevenness, which we address using a UMAP-based fuzzy graph construction.

\begin{algorithm}[!tb]
\caption{Concept Extraction}\label{alg:mental-context}
\KwData{All tokens in the input embeddings.}
\KwResult{A set of hierarchical communities.}
 Create a community list. The initial community is the entire space\;
 \For{k=[different neighbor sizes]}{
  \For{all communities}{
       {\small{- Generate knn graph from the input embedding weights\;}}
       {\small{- Compute the edge weights of the graph using fuzzy simplex\;}}
       {\small{- Apply Louvain community detection\;}}
       {\small{- Add the identified communities to the list\;}}
  }
 }
\end{algorithm}
\subsection{Graph Construction}\label{umap-knn}
The first step in the Uniform Manifold Approximation and Projection (UMAP) algorithm is to approximate the manifold by constructing a fuzzy topological representation of the embeddings using nearest neighbor descent \cite{mcinnes2020umap}.
Inspired by that, we use the same nearest neighbor descent method to find the $K$ nearest neighbors for every token embedding in the embedding space and then use the same equations used in UMAP's fuzzy graph construction to define the weight function of the edge between $x_i$ and $x_j$ nodes in the $K$-NN graph \cite{mcinnes2020umap}:
\begin{equation} 
\label{weight}
\omega(x_i,x_j) = exp \left(\frac{-\max(0,~ d(x_i,x_j) - \rho_i)}{\sigma_i}\right)
\end{equation}
where $d(\cdot,\cdot)$ is the distance function (cosine in our case) and $\rho_i$ is calculated as:
\begin{equation} 
\label{rho}
\rho_i = \min \{ d(x_i, x_{j}) \mid 1 \leq j \leq k, \; d(x_i, x_{j}) > 0\}
\end{equation}
where $k$ is the number of neighbors of node $i$. Finally, $\sigma_i$ is calculated by setting the summation of weights of a node to be equal to a constant (i.e., $\log_2(k)$):
\begin{equation} 
\label{sigma}
\sum_{j=1}^{k} \omega(x_i,x_j) = \log_2(k).
\end{equation}
Building upon our theoretical arguments, the conceptual/categorical representations (if they exist), should form fuzzy partitions that can be detected by the community detection algorithms. Note that since UMAP dimensionality reduction process can lead to information loss \cite{geiger2012relative, JMLR:v22:20-1061}, potentially obscuring important nuances in the representations, we perform community detection in the high-dimensional space. 

\subsection{Louvain Community Detection}
The Louvain method is a widely used algorithm for community detection in large networks. It finds the communities by optimizing a metric called modularity. The modularity of a partition is a scalar value between -1 and 1 that measures how much more densely connected the nodes within a community are compared to how connected they would be in a random network. \cite{blondel2008fast}. For a weighted graph, modularity is defined as:
\begin{equation} 
\label{louvain}
Q = \frac{1}{2m}\sum_{i,j}\left[A_{ij} - \frac{k_{i}k_{j}}{2m}\right]\delta(c_i,c_j)
\end{equation}
where $A_{ij}$ represents the weight of the edge between $i$ and j; $k_i$ is the sum of the weights of the edges attached to vertex $i$; $m$ is the sum of all of the edge weights in the graph; the $\delta$-function $\delta(c_{i}, c_{i})$ is 1 if $u = v$ and 0 otherwise; $c_i$ is the community to which the nodes $i$ belongs to. \\
Then, it aggregates the communities to identify possible hierarchical structures. In this phase, each community is considered as a single node and the links between the new nodes are calculated as the sum of the weight of the links between nodes in the corresponding two communities. More details are given in appendix \ref{app:louvain}, algorithm \ref{alg:louvain}.

\subsection{Concept Extraction Algorithm}
For our concept extraction algorithm, as the first step, we create and weight the adjacency graph using  $K$-NN, UMAP-based weighing formula (mentioned in section \ref{umap-knn}), and then use Louvain algorithm. 
Algorithm \ref{alg:mental-context} describes the concept extraction process (see appendix \ref{alg:consideration} for details on the algorithm methodology and considerations).
We configured our algorithms to create $k$-NN graph iteratively for different values of $k$. This enables us to observe the communities/concepts formation at various granularities. Table \ref{tab:clusters} shows the number of identified clusters for $k = [100, 75, 50, 25, 12, 6]$. 

\begin{table}[tb!]
\caption{Number of communities with different granularities of nearest neighbors for Albert, T5, Glove, and Llama3. For Glove, we only used the subset of GloVe that present in Albert vocabulary}
\centering
\resizebox{\linewidth}{!}{
\begin{tabular}{ |l|cccccc|c|}
\hline
{}&{}&{}&{$K$}&{settings}&{}&{}&{Vocab}\\
{Models}&{100}&{75}&{50}&{25}&{12}&{6}&{Size}\\\hline 
{T5}&{1}&{115}&{1203}&{4551}&{8137}&{9407}&{32000}\\\hline 
{Albert}&{8}&{133}&{1058}&{4442}&{7718}&{8626}&{30000}\\\hline 
{GloVe}&{9}&{207}&{1157}&{3521}&{6237}&{7200}&{25869}\\\hline 
{Llama-3}&{7}&{23}&{844}&{6044}&{18644}&{32535}&{128256}\\\hline 
\end{tabular}}
\label{tab:clusters}
\end{table} 
\noindent\textbf{Hierarchy Formation.} When examining the broader perspective (i.e., $k$=100), the model primarily found groups of named entities (names of people and places), adverbs, sub-words, some number symbols, and etc. (appendix \ref{app:diagram}, Figure \ref{fig:overal_structure} shows the overall concept hierarchies of the Albert vocab). 
Zooming in further (e.g., $k$=75), these communities revealed more specific clusters that are relatable to the real world. For instance, within named entities, clusters formed for personal vs. location names, even further pinpointing locations by country. 
As the granularity level increases (approaching a smaller $k$ value), clusters exhibit a stronger association with word forms. 

\begin{figure}[!tb]
\centering
\includegraphics[width=1.0\columnwidth]
{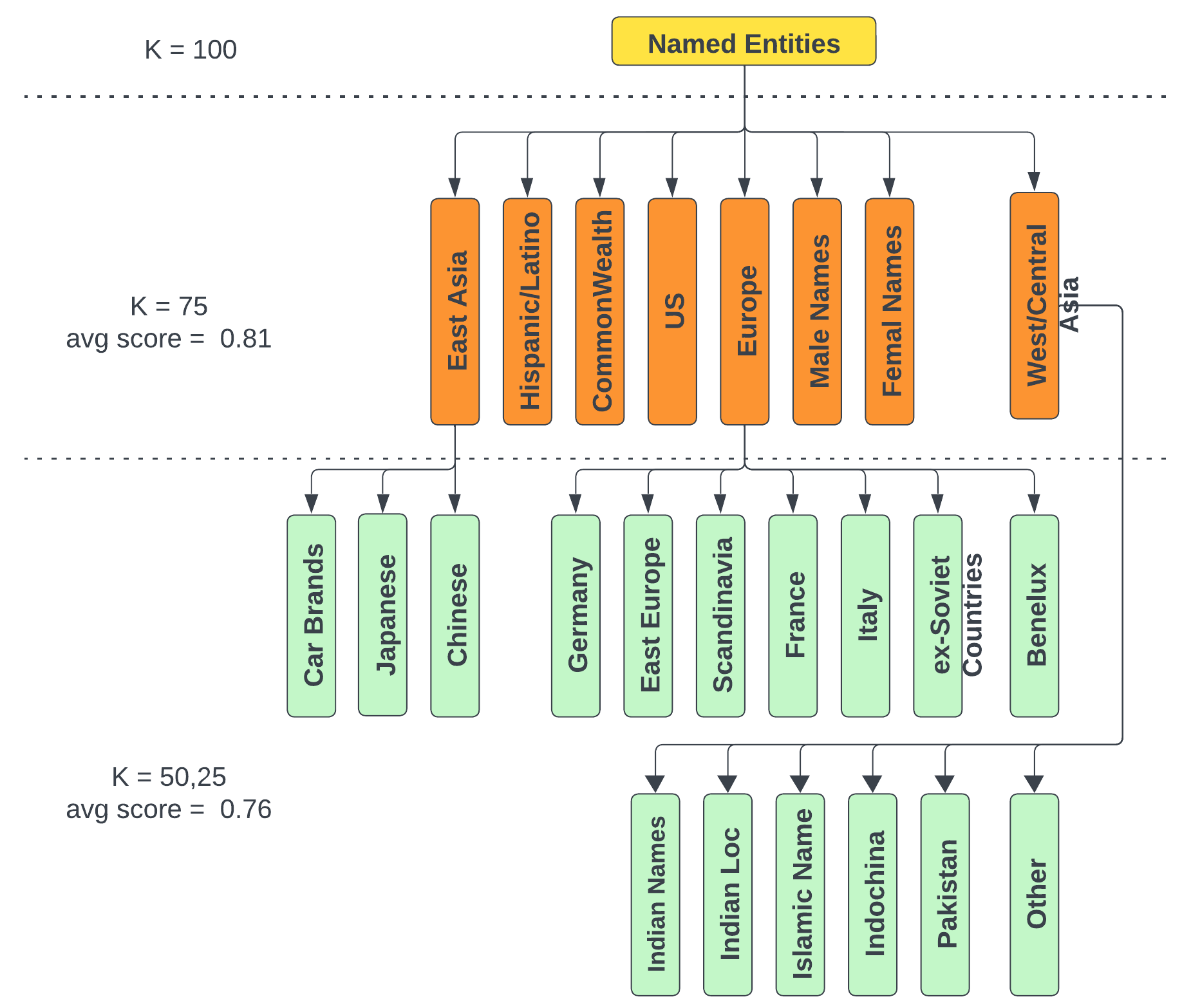}
\caption{The identified name and location communities for different k granularity for Albert model. At the left-side the average precision score for the extracted graph within each granularity is given. For the more detailed tables, see \ref{app:albert} (Note that the results for other models are also available in appendix \ref{app:alignment}).}
\vspace{-0.5em}
\label{fig:ne_clusters}
\end{figure}

\begin{figure*}[!tbh]
\centering
\includegraphics[width=1.0\textwidth]{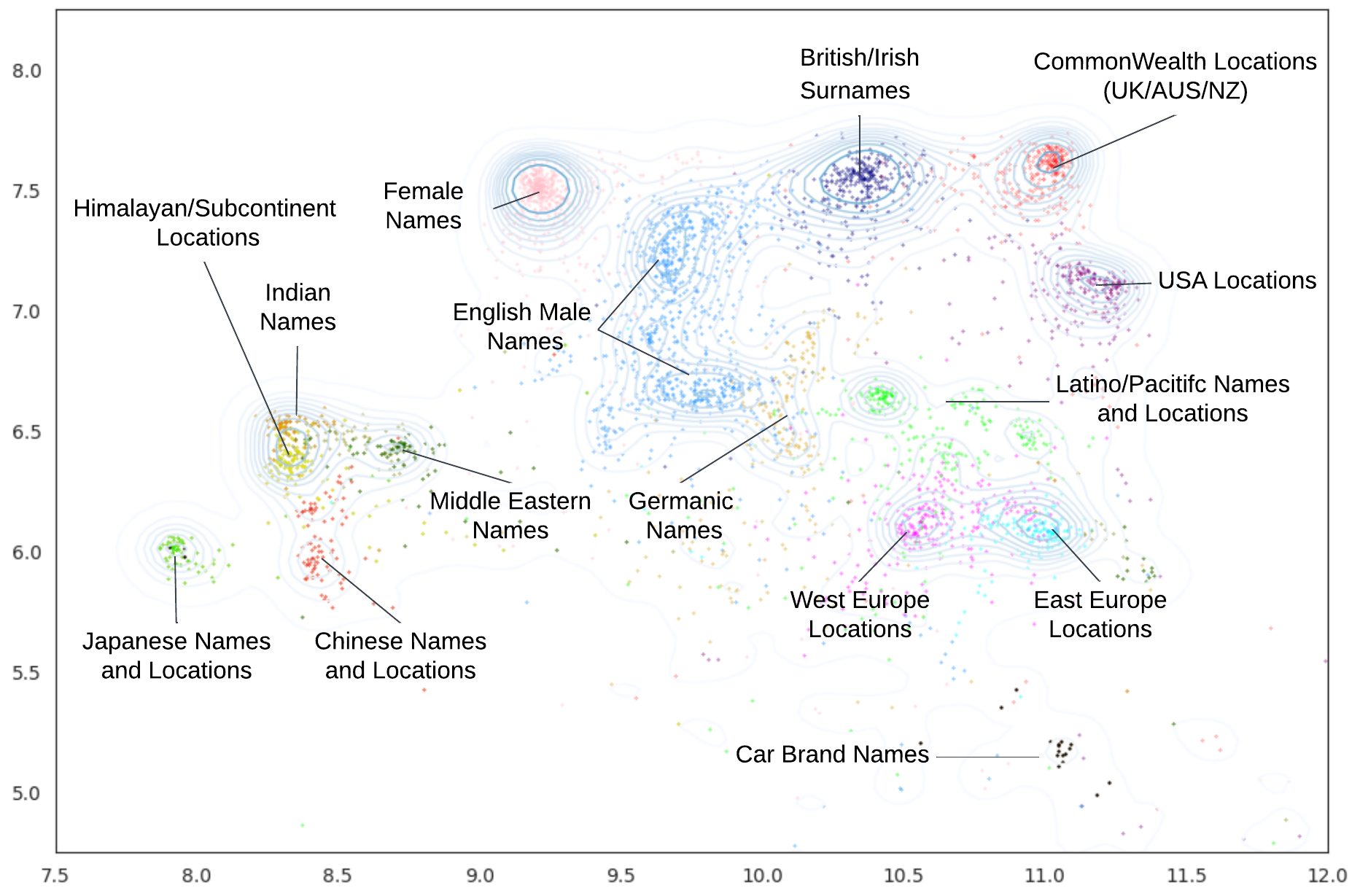}
\caption{Visualization of the identified name and location communities of size larger than 10 entities. UMAP projection along with Seaborn \cite{Waskom2021}
is used for the visualization.}
\vspace{-0.5em}
\label{fig:location_names}
\end{figure*}

\section{Evaluation: Alignment with External Knowledge}\label{human-lm-results}
We extracted several meaningful conceptual communities, ranging from symbo-numerical groups, to concrete objects and Named-Entities (such as plant groups, animals, car brands, names, and locations), to more abstract groupings such as colors, social roles and structures, currencies, etc. (more details in appendix \ref{app:diagram}).\\
To evaluate RQ1 and RQ2, we selected named entities as well as numbers, primarily because the existence of external evaluation datasets for NE as well as the well defined nature of numbers allow us to examine (1) the alignment with external world knowledge, and (2) to observe within group properties of the conceptual communities. 
To assess the consistency of conceptual groupings across different LLMs, we applied our method to the embedding spaces of GloVe \cite{pennington-etal-2014-glove}, Albert \cite{lan2019albert}, T5 \cite{raffel2020exploring}, and Llama-3 \cite{meta_llama_3} (See Appendix \ref{app:MUT}). For brevity, we present the overview of the results for Albert in the main section (for more detailed table for the all aforementioned models please refer to Appendix \ref{app:albert} and \ref{app:alignment})

\subsection{Named Entities}
To evaluate the quality and the alignment of the extracted conceptual grouping with the external world, we used we leveraged the name-dataset \cite{NameDataset2021} and the country-state-city database \cite{CountryDataset2018}. These datasets provide comprehensive information on names (including gender and country-specific popularity) and locations (including hierarchical relationships between countries, states, and cities), enabling a rigorous assessment of the conceptual groupings (see appendix \ref{app:albert} for more details).

The granularity of observed clusters was influenced by both the number of constituent samples and conceptual attributes.  For instance, English/American named entities and locations tended to form distinct clusters at higher levels of granularity (k values) due to their high frequency within the vocabulary. However, upon further increasing granularity (higher k values), hierarchical structure emerged within these clusters.  The conceptual attributes contributing to this hierarchical organization within named entities included (1) entity type (human name or location), (2) name part (first or last), gender, and (3) regional/national origin.

Note that since there's no one-to-one mapping between names/locations in LLM vocabulary and external datasets (e.g., a name may appear in multiple countries, the external dataset is also a superset), recall is less relevant. We prioritize precision to evaluate accuracy, as it better reflects our ability to identify correct matches. We observe high precision score across the identifies name entity cluster. Figure, \ref{fig:ne_clusters} shows the main groupings for Albert Model. We got average precision score of 0.81 for K=75 and 0.76 for K=50/25. Appendix \ref{app:location_names} contains the detailed tables for different models. 
Furthermore, we observed a degree of geographic ordering within the identified communities. As illustrated in Figure \ref{fig:location_names}, there appears to be a general trend from east to west as we move across the communities from bottom-left to top-right. The leftmost communities are predominantly associated with Japanese locations and names, while those on the rightmost side are primarily linked to Europe and the United States. This suggests that the model's internal representations in the input embedding layer may inherently capture geographical relationships.

\begin{table}[htb!]
\caption{Topological ordering score of different communities of the numbers. Years refer to the cluster of numbers between 1816 - 2021. Support refers to the number of samples within the cluster.}
\centering
\resizebox{0.75\linewidth}{!}{
\begin{tabular}{ |l|c|c|c|c|}
\hline
\textbf{Category}&\textbf{k-1}&\textbf{k-3}&\textbf{k-5}&\textbf{Support}\\\hline
{0-100}&{0.68}&{0.89}&{0.92}&{100}\\\hline
{100-262}&{0.75}&{0.94}&{1.0}&{115}\\\hline
{263-300}&{0.46}&{0.78}&{0.89}&{37}\\\hline
{300-400}&{0.42}&{0.71}&{0.84}&{82}\\\hline
{400+}&{0.53}&{0.75}&{0.83}&{110}\\\hline
{Years}&{0.86}&{0.96}&{0.98}&{203}\\\hline
\end{tabular}}
\label{tab:numbers}
\end{table} 

\subsection{Symbols-Numbers}
At a high level, symbols and numbers are distinctly separated from other tokens in the embedding space. Within their own domain they form communities according to (i) years, (ii) integer values, (iii) tokens indicating monetary values (e.g., \$1), (iv) ratio/time (e.g., 2:1, 3:30), (v) fractions (e.g., '/8'), (vi) large values (e.g., `100,000'), and (vii) percentages (e.g., $42\%$).
In these communities, integers create sub-communities based on their hundreds.

To align with the human notion of numbers, it is essential to identify a topological ordering within the numerical embeddings.
As the embedding manifold may not conform to a Euclidean structure \cite{law2019lorentzian, chen2021probing, cai2020isotropy}, conventional distance-based order measures were inapplicable.
Lemma \ref{topo_score} formulates the local ordering of embeddings within a embedding manifold.
\begin{lemma} Local Ordering on a Manifold: 
Let M be a manifold and let d(a, b) denote the distance between points a and b on M.  For a given positive integer $k$, we say that a point x on M is locally ordered if and only if:
\begin{equation} 
\label{topo_score}
x \in top_k(x + 1) \cap top_k(x - 1)
\end{equation} 
where $top_k(a)$ denotes the set of k-nearest neighbors of point a on M.
\end{lemma} 
\noindent This localized approach captures the intuitive concept of topological ordering within a small neighborhood, rather than relying on a global ranking.

\noindent\textbf{Topological Ordering Score} is defined as:
\begin{equation} 
\label{topo_sc2}
S = \frac{1}{n}\sum\limits_{i=0}^{n}{f_k(x_i)}
\end{equation}
where $n$ indicates the number of embeddings in the given cluster, $k$ is an integer that controls the strictness of ordering, and $f_k(x_i)$ is boolean function that returns 1 if $x_i$ hold lemma \ref{topo_score} condition.

Table \ref{tab:numbers} presents the topological ordering scores of the communities identified by our algorithm. The "k-1" score represents strict topological ordering, where $x+1$ lies exactly between $x$ and $x+2$ in the embedding space. The "k-3" and "k-5" scores measure relaxed topological ordering, where $x+1$ falls within the 3 or 5 nearest neighbors of both $x$ and $x+2$, respectively. As the results indicate, all detected communities exhibit a high degree of local topological ordering, regardless of the chosen level of strictness. 
This finding is significant because it suggests that LLMs may possess not only the ability to categorize heterogeneous input entities but also the capacity to construct meaning within smaller, internally consistent structures (internally homogeneous sub-structures). This implies the potential for LLMs to move beyond simply classifying information to actively interpreting and generating meaning within specific clusters.

\section{LLM-LLM alignment}\label{llm-llm}
Investigating LLM-LLM alignment in the input embedding layer is crucial to answer RQ3, because it reveals how well different LLMs represent the same concepts in their initial processing stages. This analysis provides valuable insights into the influence of model architecture, size, and training regimes on the formation of language representations. To quantify this alignment, we defined an alignment score that specifically captures the discrete overlaps of nearest neighbors in embedding spaces. This score offers a precise assessment of how similarly the LLMs represent shared tokens, enabling a more nuanced understanding of their representational alignment.\\
We define \textbf{Alignment Score} as:
\begin{equation} 
\label{align_score}
S = \frac{1}{n}\sum\limits_{j=0}^{j=n}\frac{|top\_k({LLM}^1_j) \cap top\_k({LLM}^2_j)|}{k}
\end{equation}
where ${LLM}^1$ and ${LLM}^2$  denote the two LLMs under investigation, $n$ represents the number of shared tokens between the LLM pair, and $top\_k({LLM}^1_j)$ and $top\_k({LLM}^2_j)$ represent the sets of the top k nearest neighbors of the $j$-th token in ${LLM}^1$ and ${LLM}^2$ respectively. The alignment score for two randomly chosen embeddings with the same vocabulary is 0 (assuming $k<<n$).

We calculated this alignment score across a range of LLMs, encompassing diverse architectures and training objectives. we first identify the set of tokens shared across the vocabularies of a given pair of LLMs. For each shared token, we compute its k nearest neighbors ($k=3,5,10,50$) in the embedding space of both models. We use pairwise cosine similarity measure to find top-k tokens within an LLM.
This selection included Albert (encoder-based), T5 (encoder-decoder), Llama 70B (decoder-only) to examine the impact of high-level model architecture. To investigate the effect of model size, we included  T5 models of different scales: T5-small, T5-base, T5-large, T5-3B, and T5-11B.
\begin{table}[tbp!]
\caption{Pair-wise alignment score for various LLM sizes, and training regimes. higher alignment score is better.}
\centering
\resizebox{1.0\linewidth}{!}{
\begin{tabular}{|c|c|cccc|c|}
\hline
{Model-1}&{Model-2}&{k=3}&{k=5}&{k=10}&{k=50}&{Support}\\\hline\hline\hline
{Albert-XXL}&{Albert-base}&{0.501}&{0.490}&{0.472}&{0.438}&{30000}\\\hline
{Albert-XXL}&{Albert-L}&{0.56}&{0.54}&{0.52}&{0.48}&{30000}\\\hline
{Albert-XXL}&{Albert-XL}&{0.60}&{0.58}&{0.55}&{0.51}&{30000}\\\hline\hline\hline
{T5-11B}&{T5-small}&{0.61}&{0.577}&{0.52}&{0.38}&{32100}\\\hline
{T5-11B}&{T5-base} &{0.68}&{0.66}&{0.60}&{0.51}&{32100}\\\hline
{T5-11B}&{T5-large}&{0.74}&{0.71}&{0.66}&{0.61}&{32100}\\\hline
{T5-11B}&{T5-3B}   &{0.79}&{0.76}&{0.71}&{0.63}&{32100}\\\hline\hline\hline
{Llama3-70B}&{Llama3-1B}&{0.57}&{0.55}&{0.50}&{0.31}&{128256}\\\hline
{Llama3-70B}&{Llama3-3B}&{0.56}&{0.53}&{0.48}&{0.27}&{128256}\\\hline\hline
{Llama3-70B}&{Llama2-70B}&{0.65}&{0.61}&{0.56}&{0.47}&{22430}\\\hline
\end{tabular}
}
\label{tab:LM-alignment}
\end{table} 
\begin{table}[tbp!]
\caption{Cross architecture alignment score for various LLM architectures, sizes, and training regimes. higher alignment score is better. Support column indicates the number of shared tokens between the LLM pair.}
\centering
\resizebox{1.02\linewidth}{!}{
\begin{tabular}{|c|c|ccccc|c|}
\hline
{Model-1}&{Model-2}&{k=3}&{k=5}&{k=10}&{k=50}&{Support}\\\hline\hline\hline
{Albert-XXL}&{Llama3-3B}&{0.55}&{0.52}&{0.49}&{0.44}&{18640}\\\hline
{Albert-XXL}&{T5-3B}&{0.63}&{0.61}&{0.59}&{0.54}&{12307}\\\hline
{Llama3-3B}&{T5-3B}&{0.66}&{0.62}&{0.58}&{0.53}&{20603}\\\hline
{Llama3-70B}&{T5-11B}&{0.64}&{0.61}&{0.56}&{0.47}&{20603}\\\hline
\end{tabular}
}
\label{tab:cross-LM-alignment}
\end{table} 
\noindent
Analyzing results from tables \ref{tab:LM-alignment} and \ref{tab:cross-LM-alignment}, we observe:
\begin{itemize}[leftmargin=*]
  \setlength{\itemsep}{0pt}  
  \setlength{\parsep}{0pt}
    \item Table \ref{tab:LM-alignment} shows that model size is a contributing factor in achieving higher alignment scores when comparing within the same architecture. Since larger models also demonstrate better performance on benchmark tasks, we can infer that the quality of the concepts formed in the input embedding layer is positively correlated with model size, assuming architecture and training regimes are held constant.
    \item Both tables \ref{tab:LM-alignment} and \ref{tab:cross-LM-alignment} show that alignment scores generally decrease as the value of k increases. This trend suggests that while models share a core understanding of semantic similarity at smaller k values, they diverge when considering more generic concepts at larger k values. The alignment drop is larger in decoder-only models, likely due to their unidirectional context \footnote{This is consistent with human studies showing that readers' eye movements are bidirectional: forward to absorb new information, and backward to resolve comprehension issues or correct errors \cite{staub2007eye}.}.
    \item Interestingly, the alignment scores between Llama2-70B and Llama3-70B are comparable to those between Llama3-70B and T5-11B (refer last rows of table \ref{tab:LM-alignment} and table \ref{tab:cross-LM-alignment}). There are minimal architectural differences between Llama2 and Llama3-70B (primarily in the context window size). This implies that factors such as training regimes (e.g., dataset size and composition) and context window size are as influential as model architecture in achieving strong alignment.
    \item Table \ref{tab:cross-LM-alignment} shows the cross-architecture alignment scores for different LLMs, comparing models of similar sizes but varying architectures. Despite architectural differences, models of similar sizes exhibit moderate to high alignment scores (mostly above 0.5), suggesting a consistent semantic organization in LLMs.
\end{itemize}

\noindent Overall, we observed moderate to high alignment across LLMs, regardless of their size, architecture, or pretraining, indicating that the findings in Section \ref{human-lm-results} may generalize to other LLMs.

\section{Bias Mitigation: Case Study of Cluster Modification}\label{app:ethnic-bias}
To investigate RQ4 and demonstrate the practical impact of our findings, we conduct a case study on modifying conceptual clusters to mitigate ethnicity bias. A key challenge here is balancing bias reduction with the preservation of the model's overall utility and linguistic integrity. Our approach focuses on embedding engineering, targeting clusters of tokens associated with stereotype-prone identities, such as proper nouns from the Indian subcontinent (e.g., Indian and Pakistani human names). We hypothesize that by modifying these token embeddings, we can disrupt learned biases without sacrificing the tokens' semantic roles within the language model. For example, we aim to investigate if the token “Sharma”, a member of the Indian proper nouns cluster, retains its identity as a proper noun after token engineering.
To evaluate fairness score, we use Bias Benchmark for Question Answering (BBQ) dataset \cite{parrish-etal-2022-bbq}, the fairness evaluation methodology is based on the WinoBias evaluation\cite{zhao2018genderbiascoreferenceresolution}. To assess the linguistic preservation, we chose the part-of-speech (POS) tagging task as a proxy indicator on CoNLL-2003 dataset \cite{conllu-2003} as well as a subset of Wikimedia dataset \cite{wikidump} (Further evaluation on the robustness of embedding engineering on GLUE, SuperGLUE, and SQUAD \cite{wang2018glue, wang2019superglue} benchmarks is provided in Appendix \ref{kg-reason-sep} to provide a more comprehensive assessment of general language model quality after the token manipulations).

Our approach begins by selecting a cluster of token embeddings from the original model. For this experiment, we used the communities associated with Indian and Pakistani human names that was identified by our concept extraction algorithm. 
Then, we calculated the mean and standard deviation of the joint cluster. Finally we modified these clusters to form a single cluster with Gaussian distribution of the joint cluster (i.e. samples from the $Gaussian(\mu, 0.7*\sigma)$). Then, we finetuned the model for POS tagging task for CoNLL-2003 dataset fro 5 epochs (see Appendix \ref{app:miti} for more details).
For the POS tagging task, we utilize the CoNLL-2003 dataset \cite{conllu-2003}, a widely-used benchmark for named entity recognition and POS tagging. Each model is fine-tuned using a standard supervised fine-tuning approach.

As shown in Table \ref{performance_fairness}, all token-engineered models exhibit over 90\% token overlap with their base counterparts. This high overlap (like the >90\% reported) suggests that even though the embeddings have been modified, they still largely represent the same underlying POS tags.  Furthermore, for those POS tags where the base and modified models show disagreement, we still observe the same overall quality. This means that even where the model does change its POS tag assignment, the new assignment is just as likely to be correct as the original. At the same time, we are able to mitigate bias ranging from 44\% for the  Albert-xxl-large and T5-3b models to 52\% for the T5-11b model. These highlight the potential of token engineering for bias mitigation while keeping the general purpose utility of the model intact.

\begin{table}[t!]
\centering
\captionsetup{justification=centering}
\caption{Fairness and POS Tagging Performance of Various Models.
Fairness: higher score implies less bias. POS: higher score implies better accuracy.}
\resizebox{1.0\columnwidth}{!}{
\begin{tabular}{|l|c|c||c|c|c|}
\hline
 & \multicolumn{2}{c|}{Fairness} & \multicolumn{3}{c|}{POS Tagging} \\ \cline{2-6}
Model & Base & Ours & Base & Our & Overlap \\
 & Score & Score & Acc & Acc &  \% \\ \hline
Albert-base & 0.26 & 0.74 & 0.91 & 0.91 & 0.90 \\ \hline
Albert-xxl & 0.28 & 0.72 & 0.93 & 0.91 & 0.92 \\ \hline
T5-3b & 0.28 & 0.72 & 0.92 & 0.91 & 0.94 \\ \hline
T5-11b & 0.24 & 0.76 & 0.94 & 0.93 & 0.94 \\ \hline
\end{tabular}
}
\label{performance_fairness}
\end{table}

\section{Conclusion}\label{conclusion}
In this paper, we propose a modular concept extraction mechanism that uncovers the emergence of distinct conceptual communities within the entire input embedding space. Using our methodology, we observe that LLMs form organized conceptual structures within their input embedding spaces. We demonstrate that the input embeddings of LLMs form categorical semantic structures that align with external world representations. We quantitatively analyze several properties of these structures, with a particular focus on categorical structures related to named entities. Additionally, we observe that numerical structures within the input embedding layer align with human notion of numerical values, including a topological ordering of numbers. We also discussed that LLMs inherently exhibit a degree of alignment with one another, suggesting the potential to extend the observed human-LLM alignment to other models. This study opens new avenues for further exploration and intervention in LLMs, especially within the realm of embedding engineering in several key areas, including bias detection and mitigation.

\newpage
\section{Limitations and Risks}\label{limitations-ref}
\noindent \textbf{Limitations:}
The model forms the conceptual communities that are meaningful but its priority is not exactly the same as that of humans. The model vocabulary is a contributing factor to the way the model prioritizes the formation of conceptual clusters in its embedding layer. For example, the number of English names is much higher than the other languages and this has caused the model to form high-level communities (e.g., $k$=75) specified for names vs. less frequent names/locations a high-level community contains the combination of regions personal and location names. This limits our method to associate the KNN resolution with the abstraction level of the extracted concepts/categories.\\
\noindent \textbf{Risks:}
This work provides detailed information about (1) the formed clusters/concept in the input embedding layer, and (2) the separation of memory from reasoning in Albert. As the methodology can also be applied to other models, it can potentially facilitate more advanced adversarial attacks and content manipulation in LLMs.

\bibliography{acl_latex}

\begin{thebibliography}{60}
\providecommand{\natexlab}[1]{#1}

\bibitem[{met(2024)}]{meta_llama_3}
 2024.
\newblock \href {https://ai.meta.com/blog/meta-llama-3/} {Introducing meta llama 3: The most capable openly available {LLM} to date}.

\bibitem[{Abdou et~al.(2021)Abdou, Kulmizev, Hershcovich, Frank, Pavlick, and S{\o}gaard}]{abdou2021can}
Mostafa Abdou, Artur Kulmizev, Daniel Hershcovich, Stella Frank, Ellie Pavlick, and Anders S{\o}gaard. 2021.
\newblock Can language models encode perceptual structure without grounding? a case study in color.
\newblock In \emph{Proceedings of the 25th Conference on Computational Natural Language Learning}, pages 109--132.

\bibitem[{Bender et~al.(2021)Bender, Gebru, McMillan-Major, and Shmitchell}]{stochastic-parrot}
Emily~M. Bender, Timnit Gebru, Angelina McMillan-Major, and Shmargaret Shmitchell. 2021.
\newblock \href {https://doi.org/10.1145/3442188.3445922} {On the dangers of stochastic parrots: Can language models be too big?}
\newblock In \emph{Proceedings of the 2021 ACM Conference on Fairness, Accountability, and Transparency}, FAccT '21, page 610–623, New York, NY, USA. Association for Computing Machinery.

\bibitem[{Bengio et~al.(2013)Bengio, Courville, and Vincent}]{bengio2013representation}
Yoshua Bengio, Aaron Courville, and Pascal Vincent. 2013.
\newblock Representation learning: A review and new perspectives.
\newblock \emph{IEEE transactions on pattern analysis and machine intelligence}, 35(8):1798--1828.

\bibitem[{Binder and Desai(2011)}]{binder2011neurobiology}
Jeffrey~R Binder and Rutvik~H Desai. 2011.
\newblock The neurobiology of semantic memory.
\newblock \emph{Trends in cognitive sciences}, 15(11):527--536.

\bibitem[{Blondel et~al.(2008)Blondel, Guillaume, Lambiotte, and Lefebvre}]{blondel2008fast}
Vincent~D Blondel, Jean-Loup Guillaume, Renaud Lambiotte, and Etienne Lefebvre. 2008.
\newblock Fast unfolding of communities in large networks.
\newblock \emph{Journal of statistical mechanics: theory and experiment}, 2008(10):P10008.

\bibitem[{Bommasani et~al.(2020)Bommasani, Davis, and Cardie}]{bommasani-etal-2020-interpreting}
Rishi Bommasani, Kelly Davis, and Claire Cardie. 2020.
\newblock \href {https://doi.org/10.18653/v1/2020.acl-main.431} {{I}nterpreting {P}retrained {C}ontextualized {R}epresentations via {R}eductions to {S}tatic {E}mbeddings}.
\newblock In \emph{Proceedings of the 58th Annual Meeting of the Association for Computational Linguistics}, pages 4758--4781, Online. Association for Computational Linguistics.

\bibitem[{Cai et~al.(2020)Cai, Huang, Bian, and Church}]{cai2020isotropy}
Xingyu Cai, Jiaji Huang, Yuchen Bian, and Kenneth Church. 2020.
\newblock Isotropy in the contextual embedding space: Clusters and manifolds.
\newblock In \emph{International conference on learning representations}.

\bibitem[{Chen et~al.(2021)Chen, Fu, Xu, Xie, Tan, Chen, and Jing}]{chen2021probing}
Boli Chen, Yao Fu, Guangwei Xu, Pengjun Xie, Chuanqi Tan, Mosha Chen, and Liping Jing. 2021.
\newblock Probing bert in hyperbolic spaces.
\newblock \emph{arXiv preprint arXiv:2104.03869}.

\bibitem[{Chiu et~al.(2016)Chiu, Korhonen, and Pyysalo}]{chiu-etal-2016-intrinsic}
Billy Chiu, Anna Korhonen, and Sampo Pyysalo. 2016.
\newblock \href {https://doi.org/10.18653/v1/W16-2501} {Intrinsic evaluation of word vectors fails to predict extrinsic performance}.
\newblock In \emph{Proceedings of the 1st Workshop on Evaluating Vector-Space Representations for {NLP}}, pages 1--6, Berlin, Germany. Association for Computational Linguistics.

\bibitem[{Devlin et~al.(2018)Devlin, Chang, Lee, and Toutanova}]{devlin2018bert}
Jacob Devlin, Ming-Wei Chang, Kenton Lee, and Kristina Toutanova. 2018.
\newblock Bert: Pre-training of deep bidirectional transformers for language understanding.
\newblock \emph{arXiv preprint arXiv:1810.04805}.

\bibitem[{Gada(2018)}]{CountryDataset2018}
Darshan Gada. 2018.
\newblock Countries states cities database.
\newblock \url{https://github.com/dr5hn/countries-states-cities-database}.

\bibitem[{Gage(1994)}]{gage1994new}
Philip Gage. 1994.
\newblock A new algorithm for data compression.
\newblock \emph{The C Users Journal}, 12(2):23--38.

\bibitem[{G{\"a}rdenfors(2020)}]{gardenfors2020primary}
Peter G{\"a}rdenfors. 2020.
\newblock Primary cognitive categories are determined by their invariances.
\newblock \emph{Frontiers in Psychology}, 11:584017.

\bibitem[{Geiger and Kubin(2012)}]{geiger2012relative}
Bernhard~C Geiger and Gernot Kubin. 2012.
\newblock Relative information loss in the pca.
\newblock In \emph{2012 IEEE information theory workshop}, pages 562--566. IEEE.

\bibitem[{Goldstein(2009)}]{goldstein2009sensation}
E.B. Goldstein. 2009.
\newblock \href {https://books.google.com/books?id=2tW91BWeNq4C} {\emph{Sensation and Perception}}.
\newblock Cengage Learning.

\bibitem[{Gu et~al.(2023)Gu, Dalvi~Mishra, and Clark}]{gu-etal-2023-language}
Yuling Gu, Bhavana Dalvi~Mishra, and Peter Clark. 2023.
\newblock \href {https://doi.org/10.18653/v1/2023.acl-long.106} {Do language models have coherent mental models of everyday things?}
\newblock In \emph{Proceedings of the 61st Annual Meeting of the Association for Computational Linguistics (Volume 1: Long Papers)}, pages 1892--1913, Toronto, Canada. Association for Computational Linguistics.

\bibitem[{Gurnani(2017)}]{gurnani-2017-hypothesis}
Nishant Gurnani. 2017.
\newblock \href {https://doi.org/10.18653/v1/W17-5303} {Hypothesis testing based intrinsic evaluation of word embeddings}.
\newblock In \emph{Proceedings of the 2nd Workshop on Evaluating Vector Space Representations for {NLP}}, pages 16--20, Copenhagen, Denmark. Association for Computational Linguistics.

\bibitem[{Gurnee and Tegmark(2024)}]{gurnee2024language}
Wes Gurnee and Max Tegmark. 2024.
\newblock \href {https://openreview.net/forum?id=jE8xbmvFin} {Language models represent space and time}.
\newblock In \emph{The Twelfth International Conference on Learning Representations}.

\bibitem[{Hampton(2007)}]{hampton2007typicality}
James~A Hampton. 2007.
\newblock Typicality, graded membership, and vagueness.
\newblock \emph{Cognitive Science}, 31(3):355--384.

\bibitem[{Ji et~al.(2023)Ji, Lee, Frieske, Yu, Su, Xu, Ishii, Bang, Madotto, and Fung}]{ji2023survey}
Ziwei Ji, Nayeon Lee, Rita Frieske, Tiezheng Yu, Dan Su, Yan Xu, Etsuko Ishii, Ye~Jin Bang, Andrea Madotto, and Pascale Fung. 2023.
\newblock Survey of hallucination in natural language generation.
\newblock \emph{ACM Computing Surveys}, 55(12):1--38.

\bibitem[{Kudo and Richardson(2018)}]{kudo-richardson-2018-sentencepiece}
Taku Kudo and John Richardson. 2018.
\newblock \href {https://doi.org/10.18653/v1/D18-2012} {{S}entence{P}iece: A simple and language independent subword tokenizer and detokenizer for neural text processing}.
\newblock In \emph{Proceedings of the 2018 Conference on Empirical Methods in Natural Language Processing: System Demonstrations}, pages 66--71, Brussels, Belgium. Association for Computational Linguistics.

\bibitem[{Lan et~al.(2019)Lan, Chen, Goodman, Gimpel, Sharma, and Soricut}]{lan2019albert}
Zhenzhong Lan, Mingda Chen, Sebastian Goodman, Kevin Gimpel, Piyush Sharma, and Radu Soricut. 2019.
\newblock Albert: A lite bert for self-supervised learning of language representations.
\newblock In \emph{International Conference on Learning Representations}.

\bibitem[{Law et~al.(2019)Law, Liao, Snell, and Zemel}]{law2019lorentzian}
Marc Law, Renjie Liao, Jake Snell, and Richard Zemel. 2019.
\newblock Lorentzian distance learning for hyperbolic representations.
\newblock In \emph{International Conference on Machine Learning}, pages 3672--3681. PMLR.

\bibitem[{Li et~al.(2021)Li, Bouraoui, Camacho-Collados, Espinosa-Anke, Gu, and Schockaert}]{ijcai2021p0530}
Na~Li, Zied Bouraoui, Jose Camacho-Collados, Luis Espinosa-Anke, Qing Gu, and Steven Schockaert. 2021.
\newblock \href {https://doi.org/10.24963/ijcai.2021/530} {Modelling general properties of nouns by selectively averaging contextualised embeddings}.
\newblock In \emph{Proceedings of the Thirtieth International Joint Conference on Artificial Intelligence, {IJCAI-21}}, pages 3850--3856. International Joint Conferences on Artificial Intelligence Organization.
\newblock Main Track.

\bibitem[{Liang et~al.(2021)Liang, Wu, Morency, and Salakhutdinov}]{pmlr-v139-liang21a}
Paul~Pu Liang, Chiyu Wu, Louis-Philippe Morency, and Ruslan Salakhutdinov. 2021.
\newblock \href {https://proceedings.mlr.press/v139/liang21a.html} {Towards understanding and mitigating social biases in language models}.
\newblock In \emph{Proceedings of the 38th International Conference on Machine Learning}, volume 139 of \emph{Proceedings of Machine Learning Research}, pages 6565--6576. PMLR.

\bibitem[{Liu et~al.(2019)Liu, Gardner, Belinkov, Peters, and Smith}]{liu-etal-2019-linguistic}
Nelson~F. Liu, Matt Gardner, Yonatan Belinkov, Matthew~E. Peters, and Noah~A. Smith. 2019.
\newblock \href {https://doi.org/10.18653/v1/N19-1112} {Linguistic knowledge and transferability of contextual representations}.
\newblock In \emph{Proceedings of the 2019 Conference of the North {A}merican Chapter of the Association for Computational Linguistics: Human Language Technologies, Volume 1 (Long and Short Papers)}, pages 1073--1094, Minneapolis, Minnesota. Association for Computational Linguistics.

\bibitem[{Love and Gureckis(2007)}]{love2007clustering}
Bradley~C Love and Todd~M Gureckis. 2007.
\newblock Models in search of a brain.
\newblock \emph{Cognitive, Affective, \& Behavioral Neuroscience}, 7(2):90--108.

\bibitem[{Lukas et~al.(2023)Lukas, Salem, Sim, Tople, Wutschitz, and Zanella-Beguelin}]{10179300}
N.~Lukas, A.~Salem, R.~Sim, S.~Tople, L.~Wutschitz, and S.~Zanella-Beguelin. 2023.
\newblock \href {https://doi.org/10.1109/SP46215.2023.10179300} {Analyzing leakage of personally identifiable information in language models}.
\newblock In \emph{2023 IEEE Symposium on Security and Privacy (SP)}, pages 346--363, Los Alamitos, CA, USA. IEEE Computer Society.

\bibitem[{Mann et~al.(2020)Mann, Ryder, Subbiah, Kaplan, Dhariwal, Neelakantan, Shyam, Sastry, Askell, Agarwal et~al.}]{mann2020language}
Ben Mann, N~Ryder, M~Subbiah, J~Kaplan, P~Dhariwal, A~Neelakantan, P~Shyam, G~Sastry, A~Askell, S~Agarwal, et~al. 2020.
\newblock Language models are few-shot learners.
\newblock \emph{arXiv preprint arXiv:2005.14165}.

\bibitem[{McCoy et~al.(2019)McCoy, Pavlick, and Linzen}]{mccoy-etal-2019-right}
Tom McCoy, Ellie Pavlick, and Tal Linzen. 2019.
\newblock \href {https://doi.org/10.18653/v1/P19-1334} {Right for the wrong reasons: Diagnosing syntactic heuristics in natural language inference}.
\newblock In \emph{Proceedings of the 57th Annual Meeting of the Association for Computational Linguistics}, pages 3428--3448, Florence, Italy. Association for Computational Linguistics.

\bibitem[{McInnes et~al.(2020)McInnes, Healy, and Melville}]{mcinnes2020umap}
Leland McInnes, John Healy, and James Melville. 2020.
\newblock \href {https://arxiv.org/abs/1802.03426} {Umap: Uniform manifold approximation and projection for dimension reduction}.
\newblock \emph{Preprint}, arXiv:1802.03426.

\bibitem[{Mikolov et~al.(2013)Mikolov, Chen, Corrado, and Dean}]{mikolov2013efficient}
Tomas Mikolov, Kai Chen, Greg Corrado, and Jeffrey Dean. 2013.
\newblock Efficient estimation of word representations in vector space.
\newblock \emph{arXiv preprint arXiv:1301.3781}.

\bibitem[{Morris et~al.(2023)Morris, Kuleshov, Shmatikov, and Rush}]{morris-etal-2023-text}
John Morris, Volodymyr Kuleshov, Vitaly Shmatikov, and Alexander Rush. 2023.
\newblock \href {https://doi.org/10.18653/v1/2023.emnlp-main.765} {Text embeddings reveal (almost) as much as text}.
\newblock In \emph{Proceedings of the 2023 Conference on Empirical Methods in Natural Language Processing}, pages 12448--12460, Singapore. Association for Computational Linguistics.

\bibitem[{Niven and Kao(2019)}]{niven-kao-2019-probing}
Timothy Niven and Hung-Yu Kao. 2019.
\newblock \href {https://doi.org/10.18653/v1/P19-1459} {Probing neural network comprehension of natural language arguments}.
\newblock In \emph{Proceedings of the 57th Annual Meeting of the Association for Computational Linguistics}, pages 4658--4664, Florence, Italy. Association for Computational Linguistics.

\bibitem[{Pan et~al.(2023)Pan, Pan, Chen, Nakov, Kan, and Wang}]{pan-etal-2023-risk}
Yikang Pan, Liangming Pan, Wenhu Chen, Preslav Nakov, Min-Yen Kan, and William Wang. 2023.
\newblock \href {https://doi.org/10.18653/v1/2023.findings-emnlp.97} {On the risk of misinformation pollution with large language models}.
\newblock In \emph{Findings of the Association for Computational Linguistics: EMNLP 2023}, pages 1389--1403, Singapore. Association for Computational Linguistics.

\bibitem[{Parrish et~al.(2022)Parrish, Chen, Nangia, Padmakumar, Phang, Thompson, Htut, and Bowman}]{parrish-etal-2022-bbq}
Alicia Parrish, Angelica Chen, Nikita Nangia, Vishakh Padmakumar, Jason Phang, Jana Thompson, Phu~Mon Htut, and Samuel Bowman. 2022.
\newblock \href {https://doi.org/10.18653/v1/2022.findings-acl.165} {{BBQ}: A hand-built bias benchmark for question answering}.
\newblock In \emph{Findings of the Association for Computational Linguistics: ACL 2022}, pages 2086--2105, Dublin, Ireland. Association for Computational Linguistics.

\bibitem[{Patel and Pavlick(2021)}]{patel2021mapping}
Roma Patel and Ellie Pavlick. 2021.
\newblock Mapping language models to grounded conceptual spaces.
\newblock In \emph{International Conference on Learning Representations}.

\bibitem[{Peng et~al.(2022)Peng, Wang, Hu, Jin, Hou, Li, Liu, and Liu}]{peng2022copen}
Hao Peng, Xiaozhi Wang, Shengding Hu, Hailong Jin, Lei Hou, Juanzi Li, Zhiyuan Liu, and Qun Liu. 2022.
\newblock Copen: Probing conceptual knowledge in pre-trained language models.
\newblock In \emph{Proceedings of the 2022 Conference on Empirical Methods in Natural Language Processing}, pages 5015--5035.

\bibitem[{Pennington et~al.(2014)Pennington, Socher, and Manning}]{pennington-etal-2014-glove}
Jeffrey Pennington, Richard Socher, and Christopher Manning. 2014.
\newblock \href {https://doi.org/10.3115/v1/D14-1162} {{G}lo{V}e: Global vectors for word representation}.
\newblock In \emph{Proceedings of the 2014 Conference on Empirical Methods in Natural Language Processing ({EMNLP})}, pages 1532--1543, Doha, Qatar. Association for Computational Linguistics.

\bibitem[{Phang et~al.(2020)Phang, Yeres, Swanson, Liu, Tenney, Htut, Vania, Wang, and Bowman}]{phang2020jiant}
Jason Phang, Phil Yeres, Jesse Swanson, Haokun Liu, Ian~F. Tenney, Phu~Mon Htut, Clara Vania, Alex Wang, and Samuel~R. Bowman. 2020.
\newblock \texttt{jiant} 2.0: A software toolkit for research on general-purpose text understanding models.
\newblock \url{http://jiant.info/}.

\bibitem[{Raffel et~al.(2020)Raffel, Shazeer, Roberts, Lee, Narang, Matena, Zhou, Li, and Liu}]{raffel2020exploring}
Colin Raffel, Noam Shazeer, Adam Roberts, Katherine Lee, Sharan Narang, Michael Matena, Yanqi Zhou, Wei Li, and Peter~J Liu. 2020.
\newblock Exploring the limits of transfer learning with a unified text-to-text transformer.
\newblock \emph{Journal of machine learning research}, 21(140):1--67.

\bibitem[{Remy(2021)}]{NameDataset2021}
Philippe Remy. 2021.
\newblock Name dataset.
\newblock \url{https://github.com/philipperemy/name-dataset}.

\bibitem[{Rogers et~al.(2021)Rogers, Kovaleva, and Rumshisky}]{rogers2021primer}
Anna Rogers, Olga Kovaleva, and Anna Rumshisky. 2021.
\newblock A primer in bertology: What we know about how bert works.
\newblock \emph{Transactions of the Association for Computational Linguistics}, 8:842--866.

\bibitem[{Sennrich et~al.(2016)Sennrich, Haddow, and Birch}]{sennrich-etal-2016-neural}
Rico Sennrich, Barry Haddow, and Alexandra Birch. 2016.
\newblock \href {https://doi.org/10.18653/v1/P16-1162} {Neural machine translation of rare words with subword units}.
\newblock In \emph{Proceedings of the 54th Annual Meeting of the Association for Computational Linguistics (Volume 1: Long Papers)}, pages 1715--1725, Berlin, Germany. Association for Computational Linguistics.

\bibitem[{Sevastjanova et~al.(2021)Sevastjanova, Kalouli, Beck, Sch{\"a}fer, and El-Assady}]{sevastjanova-etal-2021-explaining}
Rita Sevastjanova, Aikaterini-Lida Kalouli, Christin Beck, Hanna Sch{\"a}fer, and Mennatallah El-Assady. 2021.
\newblock \href {https://doi.org/10.18653/v1/2021.acl-long.39} {Explaining contextualization in language models using visual analytics}.
\newblock In \emph{Proceedings of the 59th Annual Meeting of the Association for Computational Linguistics and the 11th International Joint Conference on Natural Language Processing (Volume 1: Long Papers)}, pages 464--476, Online. Association for Computational Linguistics.

\bibitem[{Song and Raghunathan(2020)}]{song2020information}
Congzheng Song and Ananth Raghunathan. 2020.
\newblock Information leakage in embedding models.
\newblock In \emph{Proceedings of the 2020 ACM SIGSAC conference on computer and communications security}, pages 377--390.

\bibitem[{Staub and Rayner(2007)}]{staub2007eye}
Adrian Staub and Keith Rayner. 2007.
\newblock Eye movements and on-line comprehension processes.
\newblock \emph{The Oxford handbook of psycholinguistics}, 327:342.

\bibitem[{Tenney et~al.(2019)Tenney, Das, and Pavlick}]{tenney-etal-2019-bert}
Ian Tenney, Dipanjan Das, and Ellie Pavlick. 2019.
\newblock \href {https://doi.org/10.18653/v1/P19-1452} {{BERT} rediscovers the classical {NLP} pipeline}.
\newblock In \emph{Proceedings of the 57th Annual Meeting of the Association for Computational Linguistics}, pages 4593--4601, Florence, Italy. Association for Computational Linguistics.

\bibitem[{Tenney et~al.(2018)Tenney, Xia, Chen, Wang, Poliak, McCoy, Kim, Van~Durme, Bowman, Das et~al.}]{tenney2018you}
Ian Tenney, Patrick Xia, Berlin Chen, Alex Wang, Adam Poliak, R~Thomas McCoy, Najoung Kim, Benjamin Van~Durme, Samuel~R Bowman, Dipanjan Das, et~al. 2018.
\newblock What do you learn from context? probing for sentence structure in contextualized word representations.
\newblock In \emph{International Conference on Learning Representations}.

\bibitem[{Thawani et~al.(2019)Thawani, Srivastava, and Singh}]{thawani-etal-2019-swow}
Avijit Thawani, Biplav Srivastava, and Anil Singh. 2019.
\newblock \href {https://doi.org/10.18653/v1/W19-2006} {{SWOW}-8500: Word association task for intrinsic evaluation of word embeddings}.
\newblock In \emph{Proceedings of the 3rd Workshop on Evaluating Vector Space Representations for {NLP}}, pages 43--51, Minneapolis, USA. Association for Computational Linguistics.

\bibitem[{Tjong Kim~Sang and De~Meulder(2003)}]{conllu-2003}
Erik~F. Tjong Kim~Sang and Fien De~Meulder. 2003.
\newblock \href {https://aclanthology.org/W03-0419} {Introduction to the {C}o{NLL}-2003 shared task: Language-independent named entity recognition}.
\newblock In \emph{Proceedings of the Seventh Conference on Natural Language Learning at {HLT}-{NAACL} 2003}, pages 142--147.

\bibitem[{Vaswani et~al.(2017)Vaswani, Shazeer, Parmar, Uszkoreit, Jones, Gomez, Kaiser, and Polosukhin}]{vaswani2017attention}
Ashish Vaswani, Noam Shazeer, Niki Parmar, Jakob Uszkoreit, Llion Jones, Aidan~N Gomez, {\L}ukasz Kaiser, and Illia Polosukhin. 2017.
\newblock Attention is all you need.
\newblock \emph{Advances in neural information processing systems}, 30.

\bibitem[{Wang et~al.(2019)Wang, Pruksachatkun, Nangia, Singh, Michael, Hill, Levy, and Bowman}]{wang2019superglue}
Alex Wang, Yada Pruksachatkun, Nikita Nangia, Amanpreet Singh, Julian Michael, Felix Hill, Omer Levy, and Samuel Bowman. 2019.
\newblock Superglue: A stickier benchmark for general-purpose language understanding systems.
\newblock \emph{Advances in neural information processing systems}, 32.

\bibitem[{Wang et~al.(2018)Wang, Singh, Michael, Hill, Levy, and Bowman}]{wang2018glue}
Alex Wang, Amanpreet Singh, Julian Michael, Felix Hill, Omer Levy, and Samuel~R Bowman. 2018.
\newblock Glue: A multi-task benchmark and analysis platform for natural language understanding.
\newblock \emph{arXiv preprint arXiv:1804.07461}.

\bibitem[{Wang et~al.(2021)Wang, Huang, Rudin, and Shaposhnik}]{JMLR:v22:20-1061}
Yingfan Wang, Haiyang Huang, Cynthia Rudin, and Yaron Shaposhnik. 2021.
\newblock \href {http://jmlr.org/papers/v22/20-1061.html} {Understanding how dimension reduction tools work: An empirical approach to deciphering t-sne, umap, trimap, and pacmap for data visualization}.
\newblock \emph{Journal of Machine Learning Research}, 22(201):1--73.

\bibitem[{Waskom(2021)}]{Waskom2021}
Michael~L. Waskom. 2021.
\newblock \href {https://doi.org/10.21105/joss.03021} {seaborn: statistical data visualization}.
\newblock \emph{Journal of Open Source Software}, 6(60):3021.

\bibitem[{Wikimedia()}]{wikidump}
Wikimedia.
\newblock \href {https://dumps.wikimedia.org/enwiki/20231001/} {English wikipedia dump}.

\bibitem[{Zhao et~al.(2018)Zhao, Wang, Yatskar, Ordonez, and Chang}]{zhao2018genderbiascoreferenceresolution}
Jieyu Zhao, Tianlu Wang, Mark Yatskar, Vicente Ordonez, and Kai-Wei Chang. 2018.
\newblock \href {https://arxiv.org/abs/1804.06876} {Gender bias in coreference resolution: Evaluation and debiasing methods}.
\newblock \emph{Preprint}, arXiv:1804.06876.

\bibitem[{Zhu and Ghahramani(2002)}]{zhu2002learning}
Xiaojin Zhu and Zoubin Ghahramani. 2002.
\newblock Learning from labeled and unlabeled data with label propagation.
\newblock \emph{ProQuest Number: INFORMATION TO ALL USERS}.

\end{thebibliography}

\newpage
\FloatBarrier
\appendix

\begin{figure*}[!tbp]
\centering
\includegraphics[width=1.05\textwidth]{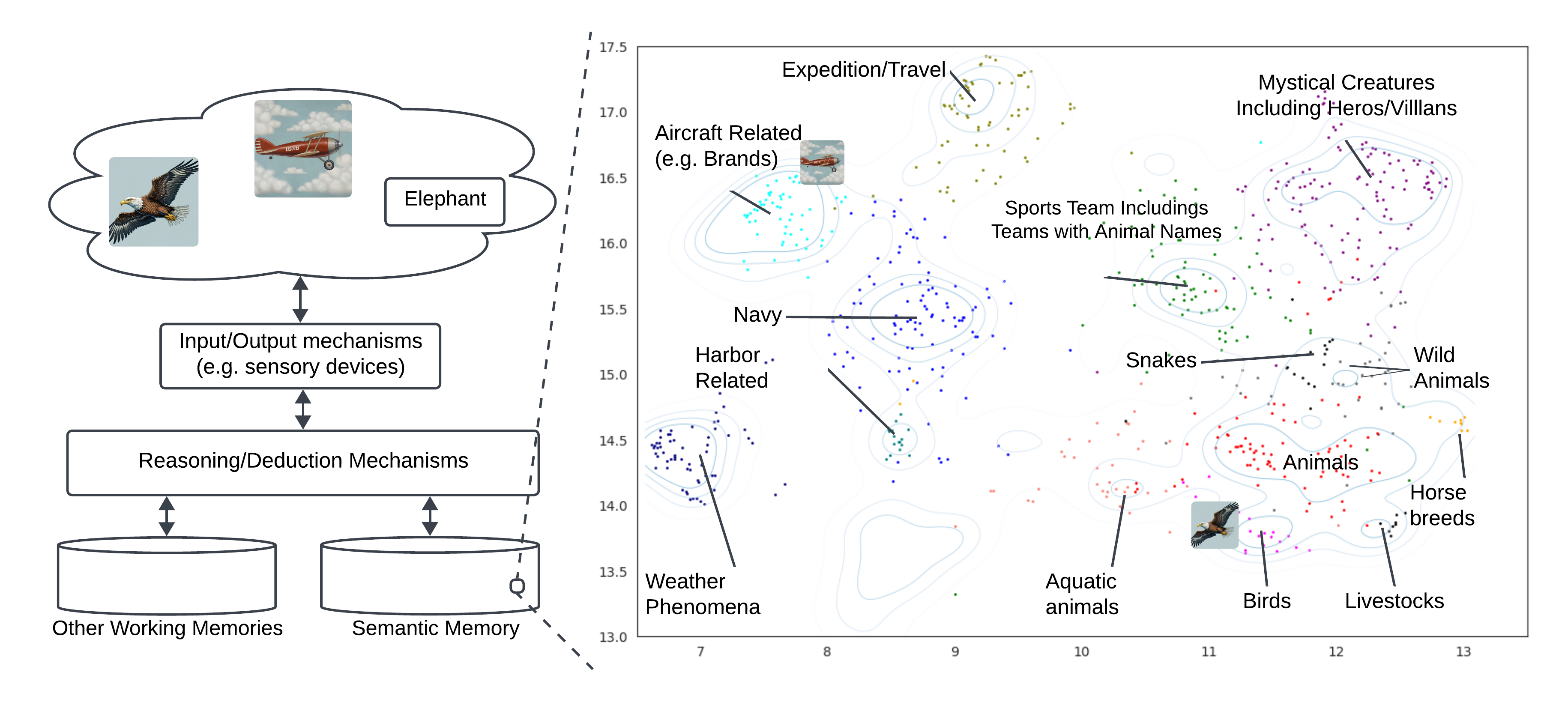}
\caption{Simplified steps on how external information is understood and retained. Upon understanding a newly encountered word/entity, it is typically stored in the semantic memory. The existence of semantic memory (on the left) allows the previously encountered words/entities to have a form of meaning even without requiring an external context. The scatter box on the right is the community (primarily associated with moving creatures) we extracted from the Albert model \cite{lan2019albert}.}
\label{fig:semantic_memory}
\end{figure*}

\section{Semantic Representation Space}
\subsection{conceptual groupings, Representation Learning and Semantic Memory}\label{concept-formation}
A simplified model of human understanding can be described as a multi-step process in which
the incoming sensory information is integrated and turned into a brain-constructed interpretation of external objects/stimuli. This results in a mental construct (or percept) \cite{goldstein2009sensation}. Then, the newly formed mental construct gets compared, integrated, and associated with the existing knowledge already retained in the semantic memory. These connections help us to conceptualize and understand the new information based on what we already know (as shown in Figure \ref{fig:semantic_memory}). 
While diverse reasoning mechanisms exist, a common thread among them seems to be their reliance on and interaction with the semantic memory. This interaction likely leaves enduring traces within such a memory \cite{binder2011neurobiology}. The integration of newly encountered mental constructs within an existing cognitive framework is often guided by their relations to established internal constructs \cite{gardenfors2020primary}. This process facilitates the creation, refinement, or expansion of semantically similar clusters \cite{love2007clustering}, implying a degree of inherent categorization\footnote{This notion implies an interrelation between the recognition and categorization.}.

\textbf{Representation Learning and Conceptual roupings)}
Representation learning is a fundamental aspect of language models, where the goal is to learn distributed representations (embeddings) for words or subword units that capture their semantic and syntactic relationships \cite{bengio2013representation}. Early models like word2vec \cite{mikolov2013efficient} utilized shallow neural networks to learn these embeddings from large text corpora. More recent models, such as Large Language Models (LLMs) like GPT-3 \cite{mann2020language}, employ transformer architectures with self-attention mechanisms \cite{vaswani2017attention}, enabling more accurate and dynamic representation learning.
From the distributional hypothesis, models can form "concepts" by identifying patterns and relationships within data, particularly through recognizing the approximate invariance of shared features across different data instances  \cite{gardenfors2020primary}. It should be noted that "conceptual groupings" is a byproduct of the learning process, not a guaranteed outcome. Furthermore, the formed "concepts" might not always align with human-defined concepts. 
Thus, it is crucial to investigate and evaluate the nature of conceptual groupings within models and their alignment with human understanding.

\subsection{Evaluation Methods}
The analysis, measurement, and interpretability of semantic representation learning in language models have been the subject of extensive research. Various methods have been proposed to evaluate how well these models capture semantic meaning. 

\textit{Intrinsic evaluation methods}, assess semantic representations by measuring word embedding similarity, comparing them to human judgments of relatedness,  (e.g. \cite{mikolov2013efficient, gurnani-2017-hypothesis, thawani-etal-2019-swow, niven-kao-2019-probing, tenney-etal-2019-bert}).  While these methods provide valuable insights, they often suffer from the limitations that probing techniques impose, and may not fully capture how well a model forms conceptual hierarchies or grasps the full spectrum of semantic relationships, as word embeddings typically represent individual words rather than higher-level categories \cite{chiu-etal-2016-intrinsic}. This limitation can hinder the evaluation of a model's ability to understand broader semantic relationships and its capacity for abstract reasoning. 

Another approach is to use \textit{extrinsic evaluation methods}, which measure the performance of language models on downstream tasks that rely on semantic understanding, such as sentiment analysis, question answering, and machine translation \cite{wang2018glue, wang2019superglue}. The performance on these tasks can indirectly indicate the quality of the learned semantic representations. The main challenge with using such extrinsic evaluation metrics to assess conceptual groupings in the embedding space is that they do not directly measure the quality of the concepts themselves.
Instead, they measure how well the model performs on specific tasks that rely on those concepts, which does not necessarily prove that the model has formed robust, human-like concepts. For instance, a model might accurately classify sentiment without truly grasping the nuances of emotions like sarcasm or irony \cite{mccoy-etal-2019-right}.

In addition to evaluation metrics, researchers have also employed visualization methods to interpret learned representations. These methods often involve projecting high-dimensional embeddings into lower-dimensional spaces for visualization \cite{sevastjanova-etal-2021-explaining, rogers2021primer}. However, several challenges arise with these approaches. Firstly, they offer indirect and subjective assessments of model understanding, lacking a quantitative basis for evaluation. Secondly, the dimensionality reduction process can lead to information loss \cite{geiger2012relative, JMLR:v22:20-1061}, potentially obscuring important nuances in the representations. Finally, different visualization techniques can produce conflicting results, making it difficult to reach definitive conclusions about the model's true comprehension. 
To mitigate these shortcomings, we propose a concept extraction mechanism that identifies communities in higher-dimensional space. This approach enables both quantitative evaluation and mitigation of potential information loss incurred during dimensionality reduction.

\section{Bias Mitigation details}\label{app:miti}
We evaluated the models on subsets of the Wikimedia dataset \cite{wikidump}, where we sample only those sentences that contain the tokens from our token-engineered cluster. For the fairness task, we adopt the Bias Benchmark for Question Answering (BBQ) dataset \cite{parrish-etal-2022-bbq}, focusing on the nationality split. We modify the data set for masked language modeling (MLM) by replacing the interrogative questions with a token '[MASK]'. To ensure the quality of evaluation, each sample is manually checked for grammatical correctness following this transformation. For evaluating fairness, we compare each base model with its corresponding token-engineered model that utilizes Gaussian-sampled embeddings. We first filter out the evaluation samples that pass the fairness test on the base model, thus isolating only the problematic cases. Our evaluation of fairness is inspired by the evaluation metrics used for the winobias dataset \cite{zhao2018genderbiascoreferenceresolution}.For the remaining biased samples, we calculate which model (base or token-engineered) is more likely to generate a biased output by examining their output logit probability. 

\section{Lovain Algorithm} \label{app:louvain}
\begin{algorithm}[!t]
\caption{Louvain}\label{alg:louvain}
\SetKw{KwGoTo}{go to}
\KwData{The initial input is a weighted network of all the nodes exist in the entire space.}
\KwResult{A set of hierarchical communities.}
\textbf{Community Detection}\label{outer_loop}\;
 Create a community list; assign a different community id to each node of the network\;
  \While{a local maxima of the modularity is \textbf{not} attained}{
    \For{each node i}{
        \For{each neighbor j}{
            evaluate the gain of modularity if i moved to the community of node j\;
            keep the maximum gain and community id\;
        }
        \If{the maximum gain is positive}
            {Move node i to the community with maximum gain.}
    }
 }
 \textbf{Community Aggregation}\;
 \If{Number of Communities > 1}{
    Reduce each community to a single node\;
     \KwGoTo \ref{outer_loop}
     }
\end{algorithm}

\noindent\textbf{Evaluation Metrics.}
conceptual groupings involves the creation of abstract internal representations through the clustering of inputs with shared features. While numerous intrinsic clustering metrics exist to assess cluster formation quality, their application to our use case is limited by two factors:
\begin{itemize}[leftmargin=*]
    \item The formed clusters are not situated within Euclidean space \cite{law2019lorentzian, chen2021probing, cai2020isotropy}, rendering geometric properties such as cluster distances inadequate indicators of concept well-formedness and distinctiveness.
    \item Concepts inherently possess a degree of vagueness \cite{hampton2007typicality}, thus metrics like compactness or separation do not reliably reflect the quality of formed concepts.
\end{itemize}

\section{Knowledge-Reasoning Separation}\label{kg-reason-sep}
Thus far, we have demonstrated that the LM constructs a knowledge base (mental representation) directly within its input embedding layer. Furthermore, we have established a degree of human-LM alignment in both hierarchical structure and semantic meaning.
\begin{table*}[tbp!]
\caption{Huggingface Albert base model on GLUE, SuperGLUE, and SQUAD tasks. For the baseline, the model was finetuned without altering the embeddings. For the mid-point, the embedding layer entries are assigned the mid-point embedding of their associated community. We used \cite{phang2020jiant} repo for benchmarking.}
    \centering
    \resizebox{0.9\textwidth}{!}{ \begin{tabular}{ |l|ccccccccc|}\hline
    \textbf{mid-point}&{}&{}&{}&{GLUE}&{}&{}&{}&{}&{}\\\hline
    \textbf{Method/Tasks}&{mnli}&{mrpc}&{qnli}&{qqp}&{rte}&{sst}&{stsb}&{wnli}&{}\\\hline
    \textbf{baseline}&{0.827}&{0.841}&{0.902}&{0.858}&{0.765}&{0.915}&{0.872}&{0.549}&{}\\\hline
    \textbf{mid-point}&{0.849}&{0.865}&{0.910}&{0.875}&{0.783}&{0.922}&{0.890}&{0.563}&{}\\\hline
    \hline
    \textbf{mid-point} &{}&{}&{}&{SuperGLUE}&{}&{}&{}&{SQUAD}&{SQUAD}\\\hline
    \textbf{Method/Tasks}&{boolq}&{cb}&{copa}&{multirc}&{record}&{wic}&{wsc} &{v1 (f1)}&{v2 (f1)}\\\hline
    \textbf{baseline} &{0.622}&{0.512}&{0.59}&{0.350}&{0.586}&{0.595}&{0.528} &{83.72}&{70.9}\\\hline
    \textbf{mid-point}& {0.621}&{0.478}&{0.55}&{0.372}&{0.588}&{0.626}&{0.634} &{84.4}&{74.9}\\\hline
    \end{tabular}}
\label{tab:glue}
\end{table*}
Now, we are interested to see the extent to which the knowledge learned during pretraining is modular and separable from the reasoning mechanisms employed by the language model in downstream tasks. For example, can the knowledge learned during the pretraining phase be selectively removed or modified without significantly impacting the model's performance on finetuning? 
The modularity can also impact the effectiveness of Language Model Inversion \cite{morris-etal-2023-text, song2020information} techniques, which aim to extract private information such as names or other sensitive details learned during the pre-training\footnote{
For example, if knowledge is found to be highly modular, it may be possible to develop targeted interventions that obscure or remove specific sensitive information without significantly impacting the model's overall performance on downstream tasks.}.\\
To investigate this, we selected GLUE, SuperGLUE, and SQUAD benchmarks as downstream tasks to assess language model performance. We then systematically removed within-community information by calculating and assigning the embedding space center (mid-point value) of each community to all its members. For example, in a community of names like "James," "John," and "Alex," all members would share the same embedding.
Table \ref{tab:glue} shows the results on major LM benchmarks GLUE, SuperGLUE, and SQUAD for the Albert base model. Although our experiment does not prove the separation of knowledge and reasoning, it indicates that at least the granular information acquired during the pretraining is not required for the model's performance on the aforementioned LM benchmarks. This is significant because it opens the door for embedding engineering of private or harmful information that is learned during the pretraining.

\section{Hierarchical Community Extraction: Methodology and Considerations}\label{alg:consideration}
There are multiple viable strategies to extract hierarchical communities in our methodology; the first one is to use algorithms such as Louvain that inherently generate hierarchical communities \cite{blondel2008fast}. 
However, the summation of the weights as a new weight in the community aggregation phase of Louvain algorithm skews the weighted graph in favor of merging smaller communities in the next phase. This detaches the community detection from the actual values in the semantic representation space (i.e. graph weights higher in the Louvain hierarchy no longer reflect the geometrical affinity of the nodes). 
Thus, for our concept extraction algorithm, we use 1-2 community aggregation and rather use KNN iteratively with different granularity for extracting hierarchical concepts (more details are described in Algorithm \ref{alg:mental-context}).

Note that we only use well-established methods such as $K$-NN, UMAP-based weighing formula, as well as Louvain  \cite{blondel2008fast} and label propagation \cite{zhu2002learning} community detection algorithms to capitalize on the established generalizability of these algorithms.  Although, as we show in the next sections that our method produces amazingly good categories, it should be noted that we intend to focus our analysis on ``if the language model forms concepts" rather than creating the most optimal concept extraction mechanism. Thus, as an extension to this work, one can focus on further optimizing our proposed method. Notably, our concept extraction is algorithm-agnostic; alternatives could be readily employed.

\section{Models Under Investigation}\label{app:MUT}
We used huggingface repository for all our models.

\subsection{Albert}
Albert (A Lite BERT) \cite{lan2019albert} is a transformer-based model for language representation learning, designed to be more efficient than its predecessor, BERT \cite{devlin2018bert}. While it shares the same basic architecture as BERT, it incorporates two main key modifications: 
\begin{itemize}[leftmargin=*]
    \item \textbf{Factorized embedding parameterization}. The benefit of factorized embedding parameterization in Albert is the significant reduction in the number of parameters compared to models like BERT. In BERT, the word embedding size (E) is tied to the hidden layer size (H), leading to a large embedding matrix as H increases. Albert instead factorizes this embedding into two smaller matrices, one projecting token ID vectors to a lower-dimensional space (E) and another projecting from this space to the hidden layer (H). This allows H to be much larger than E without increasing the parameter count of the embedding layer substantially, resulting in a more efficient use of parameters. This is particularly beneficial for large models, where memory limitations can hinder training and deployment. 
    \item \textbf{Cross-layer parameter sharing}. Parameter sharing acts as a form of regularization, preventing the model from overfitting to specific layers or features in the data. This can lead to improved generalization performance on unseen data.
\end{itemize}
Note that reducing the number of parameters and sharing information across layers can force the model to learn more general representations, thus indirectly contributing to better conceptual groupings.
\noindent\textbf{datasets.} Albert is pretrained on English text datasets, namely the English Wikipedia and BookCorpus, using self-supervised learning objectives. \\
\noindent\textbf{Tokenization.} It uses Sentencepiece tokenizer \cite{kudo-richardson-2018-sentencepiece} on the \underline{uncased} corpus with a vocabulary size limit of 30K tokens.

\subsection{T5}
T5, or Text-to-Text Transfer Transformer, is a transformer-based architecture that casts all natural language processing (NLP) tasks into a text-to-text format. This means the model takes text as input and generates text as output, regardless of the specific task. At its core, T5 is an encoder-decoder model with the following key components:
\begin{itemize}[leftmargin=*]
    \item \textbf{Encoder:} This component takes the input text and processes it into a sequence of hidden representations. It uses multiple transformer layers, each consisting of self-attention mechanisms and feedforward neural networks.
    \item \textbf{Decoder:} This component generates the output text auto-regressively, conditioned on the encoder's hidden representations. It also uses multiple transformer layers with self-attention and feedforward networks, as well as cross-attention mechanisms to attend to specific parts of the input sequence.
\end{itemize}
\noindent\textbf{Datasets.} T5 is pre-trained on a massive dataset called C4 (Colossal Clean Crawled Corpus), which contains around 750 GB of clean English text.\\
\noindent\textbf{Tokenization.} It uses Sentencepiece tokenizer on the \underline{cased} corpus with a vocabulary size limit of 30K tokens.

\subsection{GloVe}
GloVe (Global Vectors for Word Representation) is a method for obtaining vector representations for words. Unlike context-based models like transformer-based LM models, GloVe leverages global word co-occurrence statistics across a corpus to learn word vectors.
The GitHub repository\footnote{https://github.com/stanfordnlp/GloVe} provides an implementation of the GloVe model for learning word representations (word vectors or embeddings). We used the default embedding provided by the python API \footnote{Common Crawl (840B tokens, 2.2M vocab, cased, 300d vectors, 2.03 GB download), and a context window size of 10.}.\\
\noindent\textbf{Tokenization.} They used Stanford tokenizer \footnote{\url{https://nlp.stanford.edu/software/tokenizer.shtml}}, a form of BPE tokenization scheme \cite{sennrich-etal-2016-neural, gage1994new} that constructs unigram counts from a corpus, and optionally thresholds the resulting vocabulary based on total vocabulary size (2.2M \footnote{2196016 cased tokens} most frequent words in the case of GloVe embeddings) or minimum frequency count \cite{pennington-etal-2014-glove}.

\subsection{Llama3}
LLaMA 3 (Large Language Model Meta AI) is a decoder-only large language model \cite{meta_llama_3}, which Grouped Query Attention (GQA) that allows the model to effectively handle longer contexts. The model is available in various sizes, including 1B, 3, and 70B parameters. The larger models exhibit significantly improved capabilities in reasoning and complex language tasks.

\noindent\textbf{Training Dataset.} LLaMA 3 is pretrained on an extensive dataset, including over 15 trillion tokens sourced from diverse text corpora, such as books, articles, and websites. This large-scale training ensures comprehensive language understanding across different domains .


\section{Case Sensitivity Analysis}\label{case_sensitve}
The T5 model's vocabulary preserves case information, enabling us to examine how formed concepts align with capitalization differences. We identified 4,328 tokens with varying case appearances (total of 8,887 tokens). We found that for highly granular concepts (k=6), 80\% of these tokens belong to the same community. This ratio increases to 85\% for k=25 before plateauing, suggesting that case variations generally do not drastically alter the semantic grouping of tokens. This finding supports the notion that the model learns to associate words with their meanings regardless of capitalization, particularly for more abstract or broader concepts (larger k values). However, the initial increase in alignment ratio with increasing k implies that case sensitivity might still play a minor role in differentiating highly specific or nuanced concepts.


\section{ALBERT Human-Model Aligment}\label{app:albert}
\subsection{Names and Locations}\label{app:location_names}
Figure \ref{fig:location_names} visualizes the major identified communities of locations and human names. 
Generally, the formed clusters associate with specific regions or cultures and contain both location and personal names. Within these, even more granular sub-clusters emerge, characterized by distinct communities of location and personal names.
Interestingly, we observed a degree of geographic ordering within the identified communities. As illustrated in Figure \ref{fig:location_names}, there appears to be a general trend from east to west as we move across the communities from bottom-left to top-right. The leftmost communities are predominantly associated with Japanese locations and names, while those on the rightmost side are primarily linked to Europe and the United States. This suggests that the model's internal representations in the input embedding layer may inherently capture geographical relationships.

To mitigate the subjectivity risk of assessing the semantic structure, we further used external datasets in our evaluations.
For Names, we used name-dataset \cite{NameDataset2021} which consists of a comprehensive set of names (730K first names and 983K last names), their associated genders, and their popularity rank for each country.
For locations, we used the country-state-city database \cite{CountryDataset2018} which contains information on all countries, 5K+ states, and 150K+ cities. Table \ref{high_level_NE} shows the high-level communities that our approach detected. Most of the high-level communities are a mix of names/locations associated with certain geographical/cultural regions. Within these clusters, names and locations form distinct sub-communities which we discuss in more detail in the following subsections.\\
Note that since there's no one-to-one mapping between names/locations in LLM vocabulary and external datasets (e.g., a name may appear in multiple countries, the external dataset is also a superset), recall is less relevant. We prioritize precision to evaluate accuracy, as it better reflects our ability to identify correct matches.

\begin{table}[tbp!]
\caption{Precision of the largest identified name and location communities with respect to name and location databases. Note that these are at a higher level in the cluster hierarchies. Table \ref{granular_NE} shows the identified granular sub-clusters and their associated precision.}
\label{high_level_NE}
\centering
\resizebox{\linewidth}{!}{
\begin{tabular}{|l|c|c|c|}
\hline
\textbf{Category}&\textbf{Precision}&\textbf{Support}&\textbf{Note}\\\hline
{US/UK/AUS/NZ}&{0.882}&{1011}&{Human \& Location}\\\hline
{Male}&{0.854}&{946}&{Human Names}\\\hline
{Female}&{0.866}&{552}&{Human Names}\\\hline
{West-Asia}&{0.684}&{390}&{Human \& Location}\\\hline
{Hispanic/Latino}&{0.685}&{282}&{Human \& Location}\\\hline
{US}&{0.720}&{267}&{Location Names}\\\hline
{Europe}&{0.739}&{215}&{Human \& Location}\\\hline
{East-Asia}&{0.741}&{178}&{Human \& Location}\\\hline
\end{tabular}
}
\bigskip
\caption{Precision of communities based on the identified categories with respect to name-database. First and last indicates category of first and last names.}
\label{granular_NE}
\centering
\resizebox{\linewidth}{!}{
    \begin{tabular}{ |l|c|c|c|c|}
     \hline \textbf{Country}&\textbf{Overall}&\textbf{Country}&\textbf{ Sup-}&\textbf{Note}\\    \textbf{}&\textbf{Precision}&\textbf{Precision}&\textbf{port}&\textbf{}\\\hline
    {USA/UK}&{0.857}&{0.725}&{211}&{First}\\\hline
    {UK/Canada}&{0.886}&{0.698}&{116}&{First}\\\hline
    {Saudi/Arabic}&{0.82}&{0.76}&{94}&{First}\\\hline
    {Spain/Mexico}&{0.977}&{0.78}&{87}&{First}\\\hline
    {USA/Hebrew}&{0.835}&{0.568}&{81}&{First}\\\hline
    {Italy/Swiss}&{0.887}&{0.625}&{80}&{First}\\\hline
    {Belgium/France}&{0.928}&{0.789}&{76}&{First}\\\hline
    {German/Sweden}&{1.0}&{0.709}&{55}&{First}\\\hline
    {German/Austria}&{0.962}&{0.717}&{53}&{First}\\\hline
    {India}&{0.82}&{0.56}&{39}&{First}\\\hline
    {Russian}&{0.896}&{0.724}&{29}&{First}\\\hline
    \hline
    {France}&{0.9}&{0.737}&{76}&{Last}\\\hline
    {Mexico}&{0.951}&{0.855}&{83}&{Last}\\\hline
    {China}&{0.9}&{0.9}&{30}&{Last}\\\hline
    {Denmark}&{0.88}&{0.64}&{25}&{Last}\\\hline
    {Germany}&{0.95}&{0.95}&{20}&{Last}\\\hline
    {Japan}&{0.928}&{0.857}&{14}&{Last}\\\hline
    \end{tabular}
}
\bigskip
\caption{Precision of communities based on the identified categories with respect to the location database.}
\label{tab:countries}
\centering
\resizebox{0.8\linewidth}{!}{
    \begin{tabular}{ |l|c|c|}
     \hline
    \textbf{Country-Region}&\textbf{Precision}&\textbf{Support}\\\hline
    {United States}&{0.80}&{240}\\\hline
    {Germany}&{0.412}&{80}\\\hline
    {France}&{0.409}&{66}\\\hline
    {Africa}&{0.690}&{55}\\\hline
    {India}&{0.580}&{50}\\\hline
    {Italy}&{0.590}&{44}\\\hline
    {Mexico}&{0.424}&{33}\\\hline
    {Spain}&{0.592}&{27}\\\hline
    {China}&{0.500}&{20}\\\hline
    {Japan}&{0.736}&{19}\\\hline
    {Philippines}&{0.460}&{15}\\\hline
    {Pakistan}&{0.461}&{13}\\\hline
    {Netherlands}&{0.636}&{11}\\\hline
    {North-Africa}&{0.800}&{10}\\\hline
    \end{tabular}}
\end{table} 

\noindent\textbf{Names.} To determine whether the sub-clusters are associated with names, we pre-filtered the clusters that at least 70\% of their tokens are in the top 1000 names (of any country), with gender confidence of above 0.8. We primarily use gender confidence to distinguish between first-name from last-name clusters.
Then we cross-referenced all identified name communities for all countries in the dataset and assigned the country name with the highest score \footnote{For some clusters we picked the second highest country name if the scores were similar. Due to the cosmopolitan nature of countries like the USA, They tend result in a high score across the board.} as the cluster names. 

Table \ref{granular_NE} shows the precision score of the identified granular communities. The first column shows the ratio of the community members that are indeed human names (overall precision), while the second column shows the ratio with respect to specific countries. It should be noted that the country-wise scores of the countries with similar cultures and languages were similar. The reported precision score shows a high degree of categorization based on the country/culture of origin.
\footnote{It should be noted that we only included clusters with sizes larger than 10 and country-wise precision of more than 0.5 due to space limitation. The list of identified cluster names goes far beyond the aforementioned table. Clusters such as character names from books, mythology, and car brand names were also identified which were not included due to space limitations.}\\
The formation of these clusters in the input embedding space, particularly those containing ethnic minority names, presents an opportunity for token engineering to mitigate potential ethnicity biases.  (As our focus here is on interpretability and conceptual groupings and alignment, we provide an example of such a token engineering approach in Appendix \ref{app:ethnic-bias} for interested readers.)

\noindent\textbf{Locations.}
It should be noted that, although the dataset is majorly comprehensive, the LLM token space associates only one token to each location, leading to an artificial decrease in precision for multi-token location names. 
Despite these limitations, we were able to identify communities within the input embedding space that are associated not only with the location category but also with specific regions/countries. 
The precision numbers in Table \ref{tab:countries} suggest that the model groups the locations with respect to the borders of the countries, which, to a certain degree, implies a subjective perception of geographical knowledge aligned with the external world \footnote{We refer the definition of the mental map to \href{https://dictionary.apa.org/mental-map}{American Psychology Association dictionary}. It is defined as “a mental representation of the world or some part of it based on subjective perceptions rather than objective geographical knowledge.}, wherein it approximates borders and associates nationalities with specific clusters/communities. It should be noted that, due to space constraints, we only present the communities containing more than 10 entities \footnote{The complete set of clusters is included as the supplementary material.}. 
Note that dealing with multi-token location names is more challenging. For instance, "Carolina" was correctly clustered within the United States community by the LLM, while our reference dataset misclassified it as a city in Brazil. Additionally, not all location names with English spellings are included in the dataset; for example, 'Wurttemberg' (or 'Nuremberg'), a region within Germany, is absent, leading to an artificial decrease in precision.

\subsection{Social Structures}\label{social-struct}
Our methodology identifies a cluster of 903 members with the theme of social structures. In order to have a reference dataset, we annotated the dataset using a combination of GPT-4 and human annotators \footnote{
A human annotator was used alongside GPT-4 to correct GPT-4 misclassifications. Label correction by a human happens faster than the label suggestion task. We estimated our approach is more time/cost-effective while resulting in the same quality annotations.}. Given the potential human subjectivity in analyzing these concepts, we ask GPT-4 to identify the theme of each category in the datasets. (as shown in Table \ref{tab:social_structure}) for mitigating such uncertainity. In order to reduce subjectivity (we formed the class names based on the GPT-4 recommendation. Then, we asked GPT-4 and human annotator to classify each word with respect to given class names (we added another class named "Other" to avoid forcing the annotators to miscalassify).

Then, we calculated the precision score of the identified clusters with respect to our annotated dataset. We see the precision scores shown in Table \ref{tab:social_structure} as evidence that the model forms an idea on different aspects of social structure in its semantic memory. When it comes to more granular clusters ($k$=25), the sub-clusters are mostly word-forms or semantically similar words.

Intriguingly, the vocabulary model grouped words like "God" and "divinity" within the same community as concepts and structures associated with Christianity. Conversely, terms like "Islam," "Judaism," and "Talmud" formed a distinct cluster. This finding warrants further investigation to determine whether it reflects potential biases within the underlying semantic memory.

\begin{table}[tbp!]
\centering
\caption{The precision score of community members belonging to the identified categories.}
\vspace{0.3em}
\resizebox{\linewidth}{!}{
\begin{tabular}{|l|c|c|}\hline
\textbf{Category}&\textbf{Precision}&\textbf{Support}\\\hline
{Religious (Christianity)}&{0.818}&{258}\\\hline
{Military and Law Enforcement}&{0.842}&{133}\\\hline
{Administration and business}&{0.788}&{129}\\\hline
{Political Ideologies/Movements}&{0.648}&{125}\\\hline
{Monarchy and Aristocracy}&{0.64}&{107}\\\hline
{Legislature and Election}&{0.736}&{80}\\\hline
\end{tabular}}
\label{tab:social_structure}
\end{table}

\begin{figure}[!tbp]
\vspace{-1em}
\centering
\includegraphics[width=1.05\linewidth]{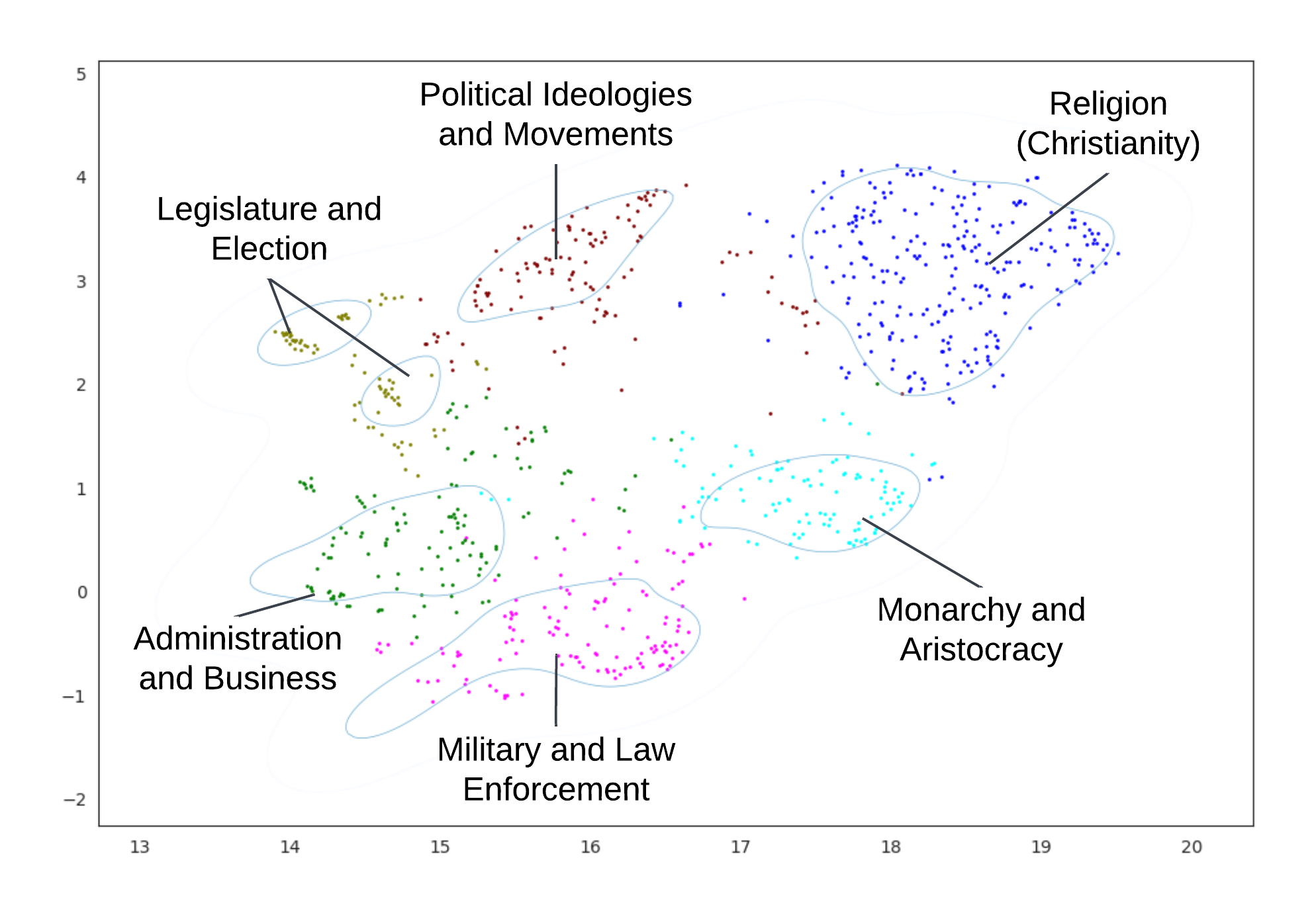}
\vspace{-1em}
\caption{Visualization of the social structure cluster and its associated identified sub-clusters.}
\label{fig:social_structure}
\end{figure}
\section{Human-LM Alignments For Glove, T5, Llama}\label{app:alignment}
\begin{table}[t!]
\centering
\caption{Location communities founds in T5 tokens. Note that most of the communities are member of 0\_0\_5, and 0\_0\_13 super communities. SOAM stands for South America.}
\resizebox{1.0\linewidth}{!}{
\begin{tabular}{|l|c|c|c|}\hline
\textbf{} & \textbf{Precision}  &\textbf{Support} &\textbf{Cluster Name}  \\\hline
USA & 0.843 & 159 & 0\_0\_5\_4   \\\hline
Britain/Ireland & 0.843 & 118 & 0\_0\_5\_5   \\\hline
Africa/SOAM & 0.672 & 64 & 0\_0\_13\_2   \\\hline
Germany & 0.619 & 21 & 0\_0\_13\_0\_0  \\\hline
Australia & 0.842 & 19 & 0\_0\_5\_10   \\\hline
France & 0.867 & 15 & 0\_0\_13\_0\_2  \\\hline
Canada & 1.000 & 13 & 0\_0\_5\_14   \\\hline
Balkan & 0.833 & 12 & 0\_0\_13\_1\_1   \\\hline
Indochina & 0.818 & 11 & 0\_0\_13\_3\_1   \\\hline
Benelux & 0.700 & 10 & 0\_0\_13\_0\_3   \\\hline
Canada & 0.778 & 9 & 0\_0\_5\_24   \\\hline
India & 0.750 & 8 & 0\_0\_13\_4\_0   \\\hline
Romania & 0.714 & 7 & 0\_0\_22\_1\_4   \\\hline
Central Europe & 0.800 & 5 & 0\_0\_13\_0\_5   \\\hline
Israel/Palestine & 0.800 & 5 & 0\_0\_13\_5\_3   \\\hline
Arab Countries & 0.800 & 5 & 0\_0\_13\_5\_5   \\\hline
Nordic & 0.750 & 4 & 0\_0\_13\_1\_6   \\\hline
Baltic & 1.000 & 4 & 0\_0\_13\_1\_8  \\\hline
\end{tabular}
}
\label{T5_locations}
\end{table}
We performed our concept extraction algorithm on Glove, Albert, and T5 models. We observed that conceptual groupings happens in all the models, however, the quality of the formed concepts is better in Albert model. 
Glove embeddings contain more than 200K tokens, in order to enable apple-to-apple comparison with Albert, we found the intersection of these tokens with the huggingface Albert-base-v2 token set, and, applied our concept extraction algorithm. We the formed clusters have weak correlation with the concepts formed in Albert model \footnote{if we assume Glove-Albert mapping exists between two clusters if more than half of their members are equivalent, \%69 percent of the Glove clusters have a corresponding Albert cluster for K=6.} (they similarities are stronger for concrete names/entities).

Tables \ref{T5_locations} and \ref{T5_names} show the T5 location and name communities detected by our algorithm. High precision numbers for these cluster indicate clear conceptual groupings. However,since the pretrained HuggingFace T5 uses cased token set for the pretraining, the number of tokens in associated with location and names are much smaller than Albert and Glove.
Table \ref{tab:llama3} shows the Llama3 name communities detected by our algorithm.  
Tables \ref{glove_locations} and \ref{glove_names} show the GloVe location and name communities detected by our algorithm. Although, it shows the categories are formed in GloVe embeddings as well, the numbers suggests the quality of the formed categories have less quality than both Albert and T5 counterpart.

\begin{table}[htb!]
\centering
\caption{Name communities founds in T5 tokens. Note that most of the communities are member of 0\_0\_5 super community.}
\resizebox{1.0\linewidth}{!}{
\begin{tabular}{|l|c|c|c|c|c|} \hline
Gender           & Overall    & Detected         & Country           & Support & Cluster      \\ 
                 & Precision  &  Country         & Precision         &         & Name         \\ \hline
Male             & 0.942      & United States    & 0.962             & 209     & 0\_0\_5\_0   \\ \hline
Female           & 0.888      & United States    & 0.858             & 162     & 0\_0\_5\_3   \\ \hline
Last Name        & 0.904      & United States    & 0.914             & 198     & 0\_0\_5\_1   \\ \hline
Male             & 0.837      & France           & 0.612             & 49      & 0\_0\_5\_7   \\ \hline
Male             & 0.909      & Peru             & 0.758             & 33      & 0\_0\_5\_8   \\ \hline
Mix              & 0.960      & Germany          & 0.84              & 25      & 0\_0\_5\_9   \\ \hline
Male             & 0.846      & Russian/Italy    & 0.615             & 13      & 0\_0\_5\_15  \\ \hline
Politicians  & 1.0        & N/A              & N/A               & 14      & 0\_0\_5\_13  \\ \hline
\end{tabular}
}
\label{T5_names}
\end{table}

\begin{table*}[htb!]
\centering
\caption{Examples of Name Entity clusters with size greater than 25 tokens found in Llama3 input embedding size. Note that we only evaluate against external datasets with English named entities.}
\label{tab:llama3}
\resizebox{0.6\linewidth}{!}{
\begin{tabular}{|l|c|c|c|c|}\hline
Type       & Precision & Support & Note            & Cluster names     \\\hline
female     & 0.757     & 393     & US/UK           & 0\_0\_0\_0\_0\_0  \\\hline
male       & 0.811     & 291     & US/UK           & 0\_0\_0\_0\_0\_2  \\\hline
male       & 0.6639    & 171     & US/UK           & 0\_0\_0\_0\_0\_3  \\\hline
male       & 0.495     & 105     & Saudi/UAE       & 0\_0\_0\_0\_0\_5  \\\hline
male       & 0.6875    & 53      & US/South Africa & 0\_0\_0\_0\_0\_11 \\\hline
male       & 0.5588    & 46      & UK/Canada       & 0\_0\_0\_0\_0\_12 \\\hline
female     & 0.571     & 41      & US/UK           & 0\_0\_0\_0\_0\_13 \\\hline
male       & 0.551     & 37      & Mexico          & 0\_0\_0\_0\_0\_14 \\\hline
male       & 0.619     & 30      & Netherlands/US  & 0\_0\_0\_0\_0\_16 \\\hline
male       & 0.666     & 30      & US/UK           & 0\_0\_0\_0\_0\_17 \\\hline
male       & 0.653     & 27      & Biblical        & 0\_0\_0\_0\_0\_18 \\\hline
last names & 0.724     & 340     & US/UK           & 0\_0\_0\_0\_0\_1  \\\hline
last names & 0.528     & 135     & Canada/US       & 0\_0\_0\_0\_0\_4  \\\hline
last names & 0.509     & 65      & UK/US           & 0\_0\_0\_0\_0\_8  \\\hline
last names & 0.717     & 39      & Mexico/Chile    & 0\_0\_0\_0\_6\_0  \\\hline
locations  & 0.75      & 485     & Mostly American  & 0\_0\_0\_8\_0   \\\hline
\end{tabular}
}
\end{table*}

\begin{table*}[htb!]
\centering
\caption{Location communities founds in the subset Glove tokens that exists in Albert Vocab.}
\resizebox{0.35\linewidth}{!}{
\begin{tabular}{|l|c|c|c|}
\hline
Country     & Precision & Support & Cluster     \\
            &           &         & Names       \\ \hline
US          & 0.674     & 522     & 0\_4        \\ \hline
UK          & 0.626     & 174     & 0\_8        \\ \hline
Europe      & 0.513     & 39      & 0\_3\_8     \\ \hline
China       & 0.625     & 32      & 0\_3\_10    \\ \hline
Italy       & 0.720     & 25      & 0\_3\_14    \\ \hline
Philippines & 0.556     & 18      & 0\_3\_15    \\ \hline
Spain       & 0.722     & 18      & 0\_3\_15    \\ \hline
Japan       & 0.688     & 16      & 0\_3\_16    \\ \hline
France      & 0.857     & 14      & 0\_3\_9\_0  \\ \hline
Africa      & 0.750     & 12      & 0\_3\_3\_1  \\ \hline
Indochina   & 0.600     & 10      & 0\_3\_11\_1 \\ \hline
Netherlands & 0.833     & 6       & 0\_3\_17\_0 \\ \hline
\end{tabular}
}
\label{glove_locations}
\end{table*}
\begin{table*}[htb!]
\centering
\caption{Name communities founds in the subset Glove tokens that exists in Albert Vocab.}
\resizebox{0.6\linewidth}{!}{
\begin{tabular}{|l|c|c|c|c|c|}
\hline
Country             & Overal    & Country   & Gender    & Support & Cluster       \\
                    & Precision & Precision &           &         & Name          \\ \hline
USA                 & 0.835     & 0.555     & female    & 575     & 0\_2\_1       \\ \hline
UK                  & 0.695     & 0.641     & male      & 223     & 0\_2\_0\_0    \\ \hline
USA                 & 0.737     & 0.337     & male      & 95      & 0\_2\_6       \\ \hline
Italy               & 0.942     & 0.692     & male      & 52      & 0\_2\_4\_1    \\ \hline
Mexico/Colombia     & 0.889     & 0.711     & male      & 45      & 0\_2\_4\_2    \\ \hline
Mexico/Peru         & 0.844     & 0.6       & male      & 45      & 0\_2\_4\_3    \\ \hline
Austria             & 0.897     & 0.793     & male      & 29      & 0\_2\_3\_3    \\ \hline
US/Nigeria         & 0.821     & 0.429     & male      & 28      & 0\_2\_0\_1\_2 \\ \hline
Russia              & 0.926     & 0.481     & male      & 27      & 0\_2\_3\_4    \\ \hline
Saudi Arabia        & 0.8       & 0.84      & male      & 25      & 0\_1\_6\_1    \\ \hline
Switzerland/Belgium & 1         & 0.846     & male      & 13      & 0\_2\_3\_0\_0 \\ \hline
UAE                 & 0.727     & 0.727     & male      & 11      & 0\_1\_6\_3    \\ \hline
Saudi-Arabia        & 0.8       & 0.7       & male      & 10      & 0\_1\_6\_0\_1 \\ \hline\hline
Germany             & 0.917     & 0.75      & no-gender & 12      & 0\_2\_3\_2\_2 \\ \hline
UK/Canada           & 0.917     & 0.33      & no-gender & 12      & 0\_2\_0\_1\_5 \\ \hline
UAE                 & 0.6       & 0.467     & no-gender & 15      & 0\_1\_6\_2    \\ \hline
\end{tabular}
}
\label{glove_names}
\end{table*}

\section{Extracted Concept Hierarchies}\label{app:diagram}
\begin{figure*}[htb!]
    \centering
    \includegraphics[width=0.66\textwidth]{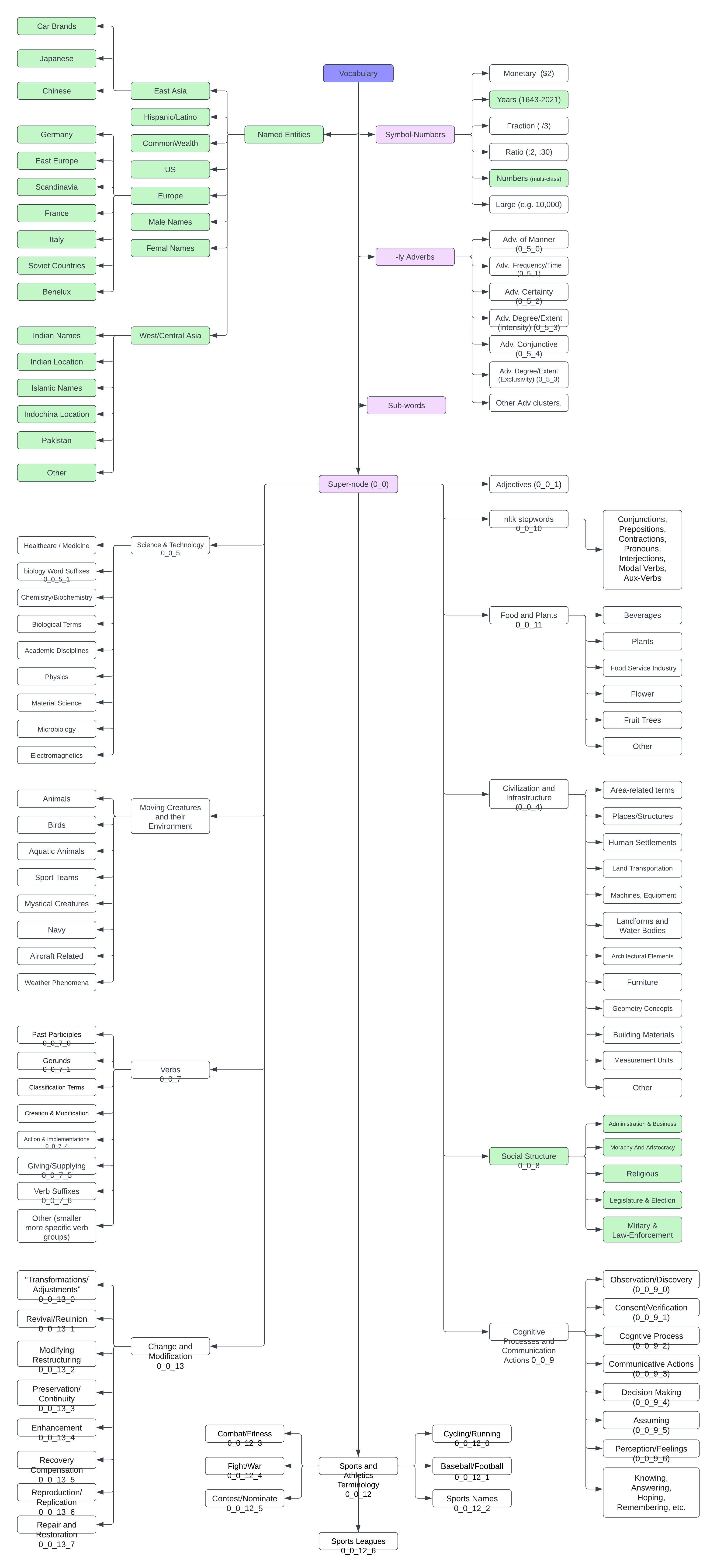}
    \vspace{-1em}
    \caption{Visualization of the hierarchical Communities from Albert. The green blocks show the clusters that being evaluated and discussed in this paper.}
    \label{fig:overal_structure}
\end{figure*}

Figure \ref{fig:overal_structure} shows the overall structure of hierarchical communities extracted by our proposed method. The cluster names were suggested by GPT-4 and corrected by a human supervisor. The green blocks are the ones that are discussed in this paper.

\end{document}